**A generalizable large-scale foundation model for musculoskeletal radiographs**

Shinn Kim[1,2†], Soobin Lee[3†], Kyoungseob Shin[3], Han-Soo Kim[1,4], Yongsung Kim[4,5], Minsu Kim[3], Juhong Nam[3], Somang Ko[6], Daeheon Kwon[3], Wook Huh[3], Ilkyu Han[1,4*], Sunghoon Kwon[2,3,7,8*]

1 Department of Orthopaedic Surgery, Seoul National University Hospital, Seoul 03080, Republic of Korea

2 Interdisciplinary Program in Bioengineering, Seoul National University, Seoul 08826, Republic of Korea

3 Department of Electrical and Computer Engineering, Seoul National University, Seoul 08826, Republic of Korea

4 Department of Orthopaedic Surgery, Seoul National University College of Medicine, Seoul 03080, Republic of Korea

5 Department of Orthopaedic Surgery, Seoul National University Bundang Hospital, Seongnam 13620, Republic of Korea

6 Department of Biomedical Science, Seoul National University College of Medicine, Seoul 03080, Republic of Korea

7 Bio-MAX Institute, Seoul National University, Seoul 08826, Republic of Korea

8 Inter-University Semiconductor Research Center, Seoul National University, Seoul 08826, Republic of Korea

† These authors contributed equally.

\* Correspondence should be addressed to I.H. (hik19@snu.ac.kr) and S.K. (skwon@snu.ac.kr)

**Abstract**

Artificial intelligence (AI) has shown promise in detecting and characterizing musculoskeletal diseases from radiographs. However, most existing models remain task-specific, annotation-dependent, and limited in generalizability across diseases and anatomical regions. Although a generalizable foundation model trained on large-scale musculoskeletal radiographs is clinically needed, publicly available datasets remain limited in size and lack sufficient diversity to enable training across a wide range of musculoskeletal conditions and anatomical sites. Here, we present SKELEX, a large-scale foundation model for musculoskeletal radiographs, trained using self-supervised learning on 1.2 million diverse, condition-rich images. The model was evaluated on 12 downstream diagnostic tasks and generally outperformed baselines in fracture detection, osteoarthritis grading, and bone tumor classification. Furthermore, SKELEX demonstrated zero-shot abnormality localization, producing error maps that identified pathologic regions without task-specific training. Building on this capability, we developed an interpretable, region-guided model for predicting bone tumors, which maintained robust performance on independent external datasets and was deployed as a publicly accessible web application. Overall, SKELEX provides a scalable, label-efficient, and generalizable AI framework for musculoskeletal imaging, establishing a foundation for both clinical translation and data-efficient research in musculoskeletal radiology.

**Introduction**

Musculoskeletal(MSK) disorders affect more than 1.7 billion people worldwide, representing a major global health challenge[1,2]. Unlike other organ systems that are dominated by a few disease categories, the MSK system encompasses a remarkably diverse spectrum of disorders, ranging from acute fractures and degenerative arthritis to inflammatory, metabolic, infectious, developmental, and malignant conditions[3]. This diversity introduces a dual layer of diagnostic complexity, as each condition may present variable manifestations across numerous anatomical regions within the musculoskeletal system. Accurate interpretation of radiographs, the first-line imaging modality for most MSK disorders, is therefore highly dependent on expert experience[4,5]. However, the global shortage of trained radiologists[6,7], combined with the increasing demand for imaging[2], is increasing diagnostic inconsistency and contributing to delays in patient care.

These diagnostic challenges underscore the need for scalable computational tools capable of assisting image interpretation across diverse diseases and anatomical regions[8]. Indeed, artificial intelligence (AI) has shown significant promise in augmenting radiographic interpretation, particularly in well-defined tasks such as traumatic fracture detection[9–12] and osteoarthritis grading[13]. Nonetheless, existing MSK AI systems remain narrowly specialized, typically trained on small, single-task datasets that rely on expert annotations[14–19]. Models designed for fracture detection often fail when encountering tumors or infections in the same image, and tools optimized for one body region rarely generalize to another. This fragmentation has hindered clinical adoption despite increasing enthusiasm for AI-assisted musculoskeletal care.

Foundation models (FMs) offer a transformative path forward. FMs are AI models trained on large, diverse datasets to learn transferable patterns rather than being confined to a single task. By learning generalizable representations from extensive and heterogeneous data, FMs address the brittleness of task-specific approaches. In other medical imaging domains, such as pathology[20,21], chest radiography[22–28], and retinal imaging[29], FMs trained on millions of images have demonstrated robust performance across numerous clinical tasks with minimal annotation. This paradigm is particularly well suited for MSK radiographs, where effective models must generalize across a wide range of conditions and anatomical regions.

Despite their potential, MSK radiographs have yet to benefit from the foundation model approach. Current publicly available datasets remain too limited in size for effective pretraining because, no single collection captures the full diversity of MSK disorders and anatomical regions. For example, the GRAZPEDWRI-DX dataset contains approximately 20,000 radiographs but focuses solely on fractures and is confined to the wrist[30]. In practice, MSK radiographs encompass more than 100 conditions, including trauma-related fractures, degenerative joint diseases, endocrine disorders, malignancies, and rare genetic syndromes, across 15 anatomical regions. Expanding beyond such narrow datasets toward large, comprehensive collections is essential for developing generalizable foundation models that can advance MSK AI toward clinical translation.

Here, we present SKELEX (musculoSKELEtal X-ray foundation model), the first large-scale foundation model for musculoskeletal radiographs, trained on the largest MSK X-ray dataset to date[31] (Fig. 1a, b). Our dataset is uniquely comprehensive, comprising more than 1.2 million radiographs that cover 15 body parts and over 89 musculoskeletal conditions. SKELEX was pretrained using a masked autoencoding strategy and subsequently adapted to multiple diagnostic applications [32] (Fig. 1c). SKELEX demonstrated strong performance across 12 musculoskeletal diagnostic tasks (Fig. 1d), even in data-sparse settings, and also achieved zero-shot abnormality localization by generating abnormality maps without task-specific training (Fig. 1e). Building on this foundation, we developed an interpretable model capable of predicting bone tumors (Fig. 1f), which we integrated into a publicly accessible web application, demonstrating the translational potential of SKELEX in clinical practice.

## Result

**Construction of musculoskeletal radiograph foundation model (SKELEX)**

To learn rich visual representations of MSK radiographs, we developed SKELEX using a two-stage self-supervised learning (SSL) pipeline. The model was initialized with a masked autoencoder (MAE) pretrained on natural images (ImageNet-1K)[32], which provided robust low-level representations. In the second stage, the pretrained backbone was further adapted to a large-scale collection of unlabeled MSK radiographs (1,296,540 images) from Seoul National University Hospital (SNUH-1M dataset), thereby enabling domain-specific representation learning across diverse anatomical regions and disease states (Supplementary Tables 1 and 2).

During pretraining, SKELEX learned to reconstruct a wide range of musculoskeletal radiographs from randomly masked inputs. This reconstruction ability generalized effectively to seven publicly available MSK radiograph datasets (Supplementary Fig. 1 and Supplementary Table 3). The model successfully reconstructed images from various body parts, preserving key anatomical landmarks (Supplementary Figs. 2 and 3). It captured both normal anatomical features and pathologic findings. For instance, when subtle radiographic clues of a bone tumor were visible, SKELEX successfully reconstructed the lesion. However, when the tumor region was completely masked and no residual cues remained, the model produced a normal-appearing reconstruction (Supplementary Fig. 4). Similar behavior was observed for other abnormalities, including fractures and osteoarthritis. These findings indicate that SKELEX acquired detailed and clinically meaningful visual representations of MSK radiographs.

**Evaluation across diverse musculoskeletal diagnostic tasks**

To assess the utility of the learned representations, we evaluated SKELEX on 12 downstream diagnostic tasks using seven publicly available datasets (Fig. 1d). Its performance was compared with two established baselines: ResNet-101[33], a convolutional neural network pretrained on ImageNet-1K using supervised learning, and ViT-L/I21K, a Vision Transformer-Large[34] model pretrained on the labeled ImageNet-21K dataset. Except for a few hyperparameters optimized for ResNet-101, all models were trained and evaluated under identical conditions (details provided in the Methods section).

SKELEX generally outperformed both baselines across a range of diagnostic tasks, including fracture detection (FracAtlas[35]), bone tumor classification (BTXRD[36]), osteoarthritis grading (OAI[37]), pes planus identification (PesPlanus dataset[38]), and other musculoskeletal conditions[30,31,39] (Methods, Supplementary Tables 4-15). Across all 12 tasks, SKELEX demonstrated average performance improvements of 2.40% over ResNet-101 and 3.89% over ViT-L/I21K, underscoring the strength and generalizability of its learned musculoskeletal representations. For instance, in bone tumor classification, SKELEX achieved an area under the receiver operating characteristic curve (AUROC) of 0.954, exceeding the performance of both ImageNet-pretrained models (ViT-L/I21K: 0.902; ResNet-101: 0.903).

Performance remained robust across anatomical regions and disease subtypes. In bone tumor classification (Fig. 2a), SKELEX maintained high accuracy across the upper limb, lower limb, and pelvis, as well as across benign and malignant tumor subtypes, indicating strong generalization beyond region- or pathology-specific features. Similarly, in fracture (Fig. 2b) and bone abnormality (Fig. 2c) classification, SKELEX achieved stable AUROC values across multiple body parts.

**Generalization across rare anatomical sites and limited labeled data**

Musculoskeletal radiographs present distinctive challenges for disease detection because of their wide anatomical and pathological diversity. In contrast, high-quality annotations which express pathological diversity are scarce. Many datasets, including MURA[31], provide only binary labels ("normal" or "abnormal"), with the latter category encompassing a broad spectrum of findings, including fractures, bone tumors, and orthopedic implants. This lack of detailed annotations and the rarity of specific disease–region combinations underscore the need for models that can generalize effectively and learn robustly from limited supervision.

SKELEX maintained strong diagnostic performance across rare anatomical regions and uncommon disease subtypes (Fig. 2a–c). In bone tumor classification, the model substantially outperformed the best comparison model in the least represented subsets, including osteofibroma (AUROC, 0.773 vs. 0.693; $P = 0.062$), and other malignant tumors (AUROC, 0.992 vs. 0.828; $P = 0.0003$). In fracture classification, SKELEX achieved superior performance for the hip region (AUROC, 1.000 vs. 0.919; $P = 0.021$), and in bone abnormality classification using MURA, it achieved higher accuracy in the least represented humerus subset (AUROC, 0.854 vs. 0.824; P = 0.012). These results indicate that SKELEX generalizes effectively under data-sparse conditions, capturing disease-relevant features that are independent of anatomical prevalence.

To further assess data efficiency, we evaluated label efficiency, defined as the amount of annotated data required to reach a target performance level, across multiple downstream tasks (Fig. 2d–f). SKELEX consistently demonstrated higher label efficiency than baseline models, requiring substantially fewer labeled samples to achieve comparable diagnostic accuracy. In bone tumor detection, SKELEX achieved an AUROC of 0.903 using only 10% of the labeled training data, matching the performance of both ResNet-101 (AUROC 0.903) and ViT-L/I21K (AUROC 0.902) trained on the full training dataset. These findings underscore the model's ability to learn effectively from limited supervision and highlight its potential as a data-efficient foundation for musculoskeletal diagnostic modeling.

**Zero-shot error map generation**

A key advantage of SKELEX lies in its ability to localize abnormalities through the generation of error maps. Because the model is trained to reconstruct masked regions of radiographs, discrepancies between reconstructed and original images reveal areas that deviate from normal anatomy, providing a self-supervised mechanism for abnormality localization without explicit lesion annotations. To generate error maps, we applied random masking to the input radiographs, reconstructed the masked regions using SKELEX, and subtracted the reconstructed outputs from the original images. Ten random masks were used per image to enhance stability, and the resulting pixel-wise difference were averaged to produce a final error map. This procedure required no additional training or fine-tuning, thus representing a zero-shot approach. Error maps were generated using the BTXRD, FracAtlas, and OAI datasets. The resulting maps highlighted regions of abnormality across bone tumor, fracture, and osteoarthritis cases (Fig. 3a–c). Quantitatively, abnormal radiographs exhibited significantly higher mean error values than normal radiographs (Fig. 3e–f), confirming the diagnostic relevance of the reconstruction error signal. Without any additional supervision, SKELEX effectively achieved zero-shot localization of abnormalities by leveraging reconstruction discrepancies.

**Region-guided interpretable framework for multi-head bone tumor classification**

Accurate interpretation of musculoskeletal radiographs requires identifying not only the disease type but also its anatomical context. Early and precise detection of primary bone tumors remains a major clinical challenge, as these lesions can mimic benign or post-traumatic findings on radiographs[40]. To enhance both diagnostic accuracy and interpretability, we extended SKELEX into a region-guided multi-head bone tumor classification framework capable of simultaneously identifying anatomical regions and multiple abnormality types, including bone tumors, fractures, and implants (Fig. 4a).

The framework consisted of two sequential stages: (1) anatomical region identification, which localized key skeletal structures, and (2) region-specific inference, where each localized region was independently analyzed for multiple abnormalities. The model was fine-tuned using three independent datasets, SNUH-BoneTumor, BTXRD[36]-Center 1 (three hospitals in China), and FracAtlas[35], to ensure broad anatomical and pathological diversity (Supplementary Table 16). To enhance generalizability across heterogeneous datasets, we implemented a label-masking strategy that computed loss only for available labels, as public datasets frequently lacked complementary annotations (for example, fracture labels in BTXRD dataset were masked).

The multi-head local classifier demonstrated strong performance in 5-fold cross-validation across all prediction tasks (Fig. 4b, c). It accurately identified anatomical regions (average AUROC = 0.998) and achieved AUROC values exceeding 0.95 for all abnormality classifications, including bone tumor, fracture, and implant detection. On the internal test set, the model exhibited strong performance across all bone tumor and fracture categories, accurately distinguishing between benign and malignant tumors as well as pathological and non-pathological fractures (Fig. 4d, e). The complete region-guided framework also maintained robust performance on independent external datasets, BTXRD-Center 2 and BTXRD-Center 3, sourced from Radiopaedia and MedPix, respectively (Fig. 4f).

Notably, the model demonstrated interpretive capability that extended beyond the labels provided in publicly available datasets. In BTXRD, which lacks explicit fracture annotations despite containing bone tumor cases with fractures, the model accurately identified both bone tumors and fractures, thereby enriching the information represented by the official dataset labels and aligning with clinical reports (Fig. 5a, Supplementary Fig. 5a). In MURA, which offers only a binary abnormality label without specifying the type of abnormality (fracture, tumor, or implant type), the model successfully inferred detailed findings, including bone tumors, fractures, implants, and specific anatomical locations (Fig. 5b, Supplementary Fig. 5b). Region-guided interpretable framework model predictions that were absent from the original dataset labels showed strong concordance with clinical notes documented by a board-certified orthopedic surgeon. To demonstrate clinical applicability, we deployed the model as a publicly accessible web-based diagnostic interface that allows users to upload radiographs and automatically estimate the probabilities of bone tumors and fractures (Fig. 6). This prototype underscores the translational potential of SKELEX-derived models for real-world musculoskeletal diagnostics.

**Discussion**

This study introduces SKELEX, a generalizable foundation model for MSK radiographs, trained on more than 1.2 million unlabeled images using a masked autoencoding strategy. SKELEX learns radiograph-specific representations directly from large-scale, uncurated clinical data, enabling the model to capture broad anatomical variation and diverse disease manifestations while minimizing the need for extensive expert annotation. The model's strong adaptability across multiple diagnostic tasks supports the notion that domain-specific self-supervised pretraining can be more effective than transferring from models trained on natural images.

Earlier musculoskeletal AI models were typically developed for a single disease category or a limited anatomical region, and many relied on small, curated datasets. Consequently, these models often lacked robustness when applied across different clinical contexts. In contrast, SKELEX was trained on data encompassing diverse anatomical regions and conditions, enabling it to perform consistently across twelve downstream diagnostic tasks and to reconstruct radiographs from multiple external datasets. This broad representational capacity is significant because musculoskeletal radiographs frequently exhibit multiple coexisting findings, such as degenerative changes, fractures, implants, and soft-tissue abnormalities. These features influence one another both visually and clinically, and accurate interpretation often depends on integrating them rather than evaluating each in isolation. Models that fail to account for this complexity may misclassify or overlook relevant findings, whereas SKELEX, through its pretraining on diverse anatomical and pathological patterns, offers a more advantageous foundation for jointly modeling these features.

A key finding of this study is the label efficiency demonstrated by SKELEX. Expert annotation in musculoskeletal imaging is time-intensive, and rare conditions are often underrepresented in existing datasets. SKELEX achieved strong performance using substantially fewer labeled examples, suggesting that large-scale self-supervised pretraining can meaningfully reduce the annotation burden required to develop clinically useful models , particularly for rare diseases where assembling large, well-annotated datasets is challenging. Furthermore, SKELEX exhibited zero-shot abnormality localization by generating reconstruction-based error maps that highlighted potentially pathological regions without additional supervision. This behavior indicates that the model internalized a representation of normal skeletal structure capable of supporting visual interpretability.

Using the pretrained model, we further developed an interpretable region-guided bone tumor detection system. This model generalized effectively to independent external datasets and demonstrated the potential to enrich public datasets that contained only binary abnormality labels by providing more detailed distinctions among tumors, fractures, and implants. These findings underscore the translational relevance of foundation models in enhancing dataset utility and supporting clinically aligned decision-making. Additionally, a web-based interface was developed to illustrate how the model could be deployed in real-world clinical and research environments.

This work carries implications for the broader goal of democratizing medical AI. By releasing SKELEX as a pretrained model, institutions with limited computational or annotation resources can adapt it to their own clinical environments using comparatively small labeled datasets. Its strong label efficiency also creates new opportunities for applying AI to rare musculoskeletal diseases, where large, well-annotated cohorts are difficult to obtain. Together, these features may help reduce disparities in access to high-quality musculoskeletal diagnostic tools across regions and healthcare systems.

Several limitations should be acknowledged. SKELEX was trained primarily on data from a single institution, which may constrain geographic and demographic diversity. We intentionally restricted training to this dataset and did not incorporate publicly available datasets to avoid any possibility of data leakage during external evaluation. Nonetheless, the model successfully reconstructed masked radiographs from seven external datasets and performed robustly across diverse diagnostic tasks, supporting meaningful generalizability. Further improvements may be achieved through federated learning or expansion to larger, multi-institutional datasets. Moreover, the current model is based on the ViT-L architecture, which constrains input resolution and may reduce sensitivity to subtle abnormalities. Employing higher-capacity transformer models or architectures optimized for high-resolution feature learning could enhance performance. Finally, this study focused exclusively on radiographs. Future research should explore multimodal approaches that integrate imaging data with clinical metadata, laboratory results, magnetic resonance imaging (MRI) or computed tomography (CT) findings, and longitudinal outcomes to enable more comprehensive diagnostic and prognostic reasoning.

In conclusion, SKELEX demonstrates that large-scale, self-supervised learning of musculoskeletal radiographs can yield a powerful and adaptable foundation model. By leveraging abundant unlabeled data, SKELEX provides a scalable framework for musculoskeletal imaging applications and presents a practical path toward more consistent, accurate, and accessible diagnostic support in clinical practice.

**Methods**

**Large-scale visual pretraining of SKELEX**

This study was approved by the Institutional Review Board of Seoul National University Hospital (IRB No. H-2203-079-1306). All radiographs and associated metadata were retrospectively collected and de-identified prior to analysis. Because this study utilized archival imaging data without direct patient contact, the requirement for informed consent was waived.

To pretrain SKELEX, we assembled a large-scale musculoskeletal radiograph dataset (SNUH-1M) for self-supervised learning. SNUH-1M comprised of 1,296,540 unlabeled musculoskeletal radiographs retrieved from the hospital's PACS between 2010 and 2016. These radiographs covered 15 anatomical regions and included a broad range of traumatic, neoplastic, degenerative, inflammatory, and postoperative conditions, thereby reflecting real-world clinical diversity (Supplementary Tables 1, 2).

SKELEX was pretrained using a masked autoencoder strategy[32]. Each radiograph was divided into patches, a random subset of which was masked, and the model was trained to reconstruct the missing patches from the visible context. This reconstruction objective encouraged the model to learn contextual and structural priors in musculoskeletal radiographs, including cortical continuity, trabecular pattern, joint alignment, and soft-tissue contours (Supplementary Figs. 1 and 2).

Pretraining settings were based on established masked autoencoder frameworks[32], with several modifications specifically tailored for radiographic data (Supplementary Table 17). The encoder was initialized with ViT-MAE ImageNet-1K pretrained weights from the Hugging Face Transformers library, using a ViT-Large backbone. The AdamW optimizer was employed with $β_2 = 0.999$ to stabilize low-level features inherited from the ImageNet initialization. Normalized pixel loss was disabled to avoid distortion from the large, uniform black background regions inherent to radiographs. A masking ratio of 75% was applied. Data augmentations included random resized cropping (20-100% of the image area) and horizontal flipping. No manual labels or anatomical supervision were used during pretraining.

**Downstream task evaluation**

SKELEX was evaluated across 12 downstream diagnostic tasks using seven publicly available musculoskeletal imaging datasets encompassing fracture detection, bone tumor classification, pes planus identification, abnormality classification, osteoarthritis grading, and bone age estimation (Supplementary Tables 4-15).

For each dataset, a held-out test set (10%) with no patient overlap was created, and the remaining data were used for five-fold cross-validation (stratified for classification tasks). The model checkpoint achieving the best validation performance, AUROC for classification tasks and MAE for regression tasks, was selected and subsequently evaluated on the held-out test set.

SKELEX was benchmarked against ViT-L/I21K and ResNet-101, both pretrained on ImageNet, using identical training, validation, and test splits. SKELEX and ViT-L/I21K were fine-tuned using the same training pipeline, whereas ResNet-101 was fine-tuned using a configuration adapted for convolutional architectures. Full hyperparameter settings are listed in Supplementary Table 18 (SKELEX and ViT-L/I21K) and Supplementary Table 19 (ResNet-101).

**Task-specific downstream evaluations**

**Pediatric wrist fracture AO classification task (GRAZPEDWRI-DX)**

A multi-class classification task was constructed to predict AO pediatric distal forearm fracture categories directly from wrist radiographs. Each image was paired with the AO classification code from the metadata column ao_classification. The ten most frequent AO categories were retained (Supplementary Table 20), and AO strings were mapped to integer indices, which served as ground-truth labels. The task was treated as single-label multi-class classification and optimized using cross-entropy loss. To avoid patient leakage, we first split the dataset into a train/validation pool and a held-out test set using a stratified patient-level split whenever a patient_id column was available in the metadata.

**Pediatric wrist fracture detection task (GRAZPEDWRI-DX)**

A binary classification task was defined to detect the presence of a radiographically visible distal forearm fracture. Radiographs of initial exams were only included and case with uncertain diagnosis (diagnosis_uncertain = 1) or cast application (cast = 1) were excluded. Radiographs in which a fracture was visible on an alternate projection but annotated as no visible fracture were also excluded to prevent label inconsistency. That made 9,924 cases with 4,472 fractured cases. We framed the task as single-label binary classification problem using binary cross-entropy loss with a sigmoid output layer.

**General fracture detection task (FracAtlas)**

Fracture detection was evaluated using the FracAtlas dataset, which contains extremity trauma radiographs labeled for the presence or absence of fracture. Metadata provided a binary fractured label. Of 4,083 images, 717 were fracture-positive. The task was formulated as binary single-label classification using binary cross-entropy loss with a sigmoid output layer.

**Abnormality classification (MURA)**

The MURA dataset contains upper-extremity radiographs labeled for radiographic abnormality. Metadata supplied a binary abnormality label, with 16,403 abnormal cases among 40,005 studies. The task was formulated as binary single-label classification with a binary cross-entropy loss with a sigmoid output layer. To prevent leakage, test dataset partitioning was performed at the patient level using the provided patient identifiers.

**Bone tumor subtyping (BTXRD)**

Fine-grained radiographic tumor subtype classification was performed using BTXRD metadata containing nine subtype labels: osteochondroma, multiple osteochondromas, simple bone cyst, giant cell tumor, osteofibroma, synovial osteochondroma, other benign tumor, osteosarcoma, and other malignant tumor. Among 3,746 radiographs, 1,867 contained a bone tumor with one of the nine subtype annotations. The task was formulated as nine-class single-label classification with cross-entropy loss. Dataset splitting occurred at the image level due to the absence of patient identifiers.

**Bone tumor benign/malignant classification (BTXRD)**

Accurately distinguishing malignant from benign bone tumors on radiographs is a clinically critical task, as early identification of malignancy directly influences the urgency of referral, need for advanced imaging or biopsy, and overall treatment strategy. To assess the ability to differentiate malignant from benign bone tumors, a binary classification task was constructed using the BTXRD malignant, benign labels. Among 1,867 tumor cases, 342 were malignant. The task was formulated as binary single-label classification using binary cross-entropy loss with a sigmoid output layer. Because patient identifiers were not available, dataset splitting was performed at the image level.

**Bone tumor presence classification (BTXRD)**

Accurate identification of bone tumors on radiographs is clinically important, as radiographs are often the first-line imaging modality for musculoskeletal complaints. Bone tumor detection was evaluated as a binary classification task using BTXRD metadata indicating tumor presence or absence (1,867 tumor cases; 1,879 non-tumor cases). The task was implemented as single-label binary classification using binary cross-entropy loss with a sigmoid output layer. Dataset splitting was performed at the image level.

**Knee osteoarthritis grading (OAI)**

Radiographic knee osteoarthritis severity was assessed using the OAI dataset, which provides knee AP radiographs matched with label of the Kellgren–Lawrence (KL) grading system. A total 8,260 knee AP radiographs were labeled 0 through 4, each corresponding to a KL grade. The task was defined as a five-class single-label classification problem optimized using cross-entropy loss. Because patient identifiers were not available, dataset partitioning was performed at the image level.

**Bone age regression (RSNA Bone Age)**

Bone age estimation was performed using the RSNA dataset of pediatric hand radiographs paired with ground-truth chronological bone age. The task was formulated as a single-target regression problem optimized with mean squared error loss. Training was performed using AdamW with a base learning rate of $1\times10^{-4}$ and weight decay 0.05; a higher learning rate ($5\times10^{-4}$) and lower weight decay ($1\times10^{-4}$) were used for ResNet-101.

**Pes Planus (flatfoot) classification (PesPlanus dataset)**

Pes planus is a condition characterized by decreased or absent medial longitudinal arch height, often identified on foot radiographs. Pes planus classification was performed using lateral foot radiographs labeled for the presence of flatfoot. Of 842 radiographs, 402 were positive. The task was defined as binary single-label classification using binary cross-entropy loss with a sigmoid output layer. Due to the absence of patient identifiers, splitting was conducted at the image level.

**Detection of orthopedic implant in pediatric wrist radiographs (GRAZPEDWRI-DX)**

To evaluate model performance in identifying the presence of orthopedic implant in pediatric wrist radiographs, we constructed the GRAZPEDWRI-DX dataset as a binary single-label classification task. Orthopedic implants commonly appear in pediatric wrist imaging following fracture management. Accurate identification of such hardware is important for downstream diagnostic tasks, as implants can obscure fracture lines, alter image characteristics, and confound automated feature extraction if not properly recognized. Among 20,327 radiographs, 708 contained implants. Models were fine-tuned end-to-end with a binary classification head using binary cross-entropy loss with a sigmoid output layer.

**Detection of orthopedic implant in general fracture dataset (FracAtlas)**

Implant detection was also evaluated using the FracAtlas dataset. Among 4,083 radiographs, 99 had implant-positive labels. The absence of patient-level identifiers required dataset splitting at the image level. This task was formulated as a binary single-label classification problem, using binary cross-entropy loss with a sigmoid output layer.

**Label efficiency**

To evaluate label efficiency, we examined how model performance varied with reductions in the number of labeled training examples. For each task, we randomly sampled 10%, 20%, 50%, or 90% of the original training set, and the remaining samples were reserved as the held-out test set. Sampling was stratified to preserve class balance across data splits. Label efficiency was evaluated across three representative tasks spanning distinct musculoskeletal clinical contexts: abnormality classification (MURA), fracture detection (FracAtlas), and bone tumor classification (BTXRD). For each subset size, SKELEX was fine-tuned using the same optimization settings as in the full-data experiments, and the checkpoint yielding the highest validation AUROC was evaluated on the corresponding test set. The same procedure was applied to ViT-L/I21K and ResNet-101 to ensure a fair comparison under equivalent training and evaluation conditions.

**Zero-shot abnormality map**

Because SKELEX was pretrained using a masked autoencoder objective, it learned to reconstruct missing image regions based on the surrounding context. When an abnormal region was entirely masked, the model often reconstructed it as anatomically normal, reflecting its learned prior of typical skeletal structures. This property enables zero-shot abnormality localization by comparing the original image with its reconstructed counterpart.

For each radiograph, we applied ten random masking patterns, reconstructed the masked image using SKELEX, and computed the per-pixel reconstruction error. Let $x \in R^{H \times W \times C}$ denote the original image and $\hat{x}^{(p)}$ the reconstructed image under the $p$-th masking pattern. Let $M^{(p)} \in \{0,1\}^{H \times W}$ denote the binary mask for pass $p$, where 1 indicates a masked pixel. The per-pixel reconstruction error for each pass is defined as: $\delta^{(p)} = (\hat{x}^{(p)} - x) \odot M^{(p)}$, where $\odot$ indicates element-wise multiplication. The per pixel squared error is calculated as

$$e^{(p)}(i,j) = \frac{1}{C}\sum_{c=1}^{C}\left(\delta_{i,j,c}^{(p)}\right)^2.$$ Accumulating over multiple random masks, the final error map is computed as the normalized average over all instances in which a pixel was masked:

$$E(i,j) = \frac{\sum_{p=1}^{n_{passes}} e^{(p)}(i,j)}{\sum_{p=1}^{n_{passes}} M^{(p)}(i,j)} \text{ for pixels with } \sum_{p=1}^{n_{passes}} M^{(p)}(i,j) > 0$$

The resulting map $E(i,j)$ highlights regions where the reconstruction deviates from normal anatomy, thereby indicating potential abnormalities. No fine-tuning or task-specific supervision was used during this process.

Zero-shot error maps were generated for three datasets (BTXRD, FRACATLAS, and OAI) to evaluate this capability across bone tumor, fracture, and osteoarthritis cases. For each dataset, randomly sampled normal radiographs were compared with abnormal radiographs. For BTXRD, radiographs frequently contained non-anatomical artifacts such as text or metallic objects from clothing or accessories. To prevent these from influencing the error maps, we cropped the anatomical region containing the tumor and compared it directly with randomly sampled normal bone regions matched by anatomical location.

**Interpretable, region guided multi-headed bone tumor classification**

We developed a two-stage framework to enable interpretable detection of bone tumors, fractures, and orthopedic implants. The first stage localized anatomically meaningful regions, and the second stage performed multi-headed classification within each region. Anatomical regions were identified using a modern real-time object detector (YOLO11x)[41]. The detector was trained on 4,049 radiographs from our institutional dataset, annotated with bounding boxes for 29 anatomical regions and auxiliary abnormal bone regions. The bounding box annotations adhered to AO Foundation's communication system for defining anatomical locations[42]. For regional classification, SKELEX was fine-tuned using a multi-headed architecture consisted of five clinically relevant output heads: (1) global abnormality detection (binary); (2) bone tumor subtype classification (four classes: malignant, intermediate, benign, normal); (3) anatomical location classification (29 classes); (4) fracture category classification (three classes: neoplastic pathologic fracture, non-neoplastic fracture, normal); (5) orthopedic implant presence detection (binary). We adopt a multi-head classification design, implemented as a single shared output tensor partitioned into task-specific groups. Supervised fine-tuning was performed using a masked multi-task objective to accommodate datasets with incomplete label coverage. For each image and label $c$, a binary label ($y_c$) and a mask ($m_c$) were defined, where ($m_c = 1$) indicated that the label was unknown or not applicable. Loss and evaluation terms were computed only for labels where ($m_c = 0$). Binary tasks were optimized using masked binary cross-entropy, and multiclass tasks were optimized using masked categorical cross-entropy. The total loss was calculated as the equally weighted sum of all active heads. Evaluation metrics (AUROC, balanced accuracy, precision, recall, and F1-score) were calculated only over unmasked labels, and grouped confusion matrices were reported for multiclass heads.

Training data for the multi-headed classification model were derived from the SNUH-BoneTumor dataset, which comprised 19,757 anatomically cropped radiographs, combined with the BTXRD-Center 1 and FracAtlas datasets. The SNUH-BoneTumor dataset was curated with detailed image-level annotations assigned by a board-certified orthopedic oncologist. When available, pathology reports from surgical specimens were used as the definitive reference standard for tumor labels. In cases without histopathologic confirmation, diagnoses were established based on radiographic findings and clinical history, supplemented by CT or MRI when available. This process ensured consistent labeling quality across benign, malignant, and non-tumorous bone lesions.

All datasets were concatenated and evaluated using a 5-fold cross-validation strategy, with an independent 10% hold-out test set reserved for final performance assessment. Images were augmented using random intersection-over-region (IoR)-constrained cropping, ±10° rotation, color jitter, and resizing to 224 × 224 pixels, followed by standard normalization. The model was initialized from SKELEX and optimized using the AdamW optimizer (learning rate = $5\times10^{-5}$, weight decay 0.02) under a cosine decay schedule with 10% warm-up over 30 epochs and a batch size of 64, employing mixed-precision training.

External evaluation was performed on two independent datasets not used during training: BTXRD-Center 2 (Radiopaedia) and BTXRD-Center 3 (MedPix). For each radiograph, the region proposal network first identified candidate anatomical regions, and each extracted region, along with the whole image, was then processed by the multi-headed classifier. Image-level predictions were derived from the aggregated regional outputs: if any region was classified as bone tumor, the entire radiograph was labeled as tumor-positive. Among tumor-positive cases, malignant bone tumor was assigned when the maximum malignant-class probability exceeded 0.5; otherwise, the case was labeled benign. Confusion matrices summarizing model performance were generated.

A web-based inference interface was implemented using FastAPI to facilitate clinical and research applications of the region-guided bone tumor classifier.

**Computing hardware and software**

To pretrain SKELEX using a masked autoencoder framework, we implemented the encoder–decoder architecture with the Hugging Face Transformers library and trained the model on a single NVIDIA RTX A6000 Ada (40 GB) GPU using CUDA 12.4 and PyTorch v2.2.2. All downstream experiments were conducted on either a single NVIDIA V100 (40 GB) GPU or an NVIDIA RTX 4090 (24 GB) GPU running CUDA 11.8 and PyTorch v2.1.2.

Checkpoints for ResNet-101 and ViT-Large, both pretrained on ImageNet, were obtained from publicly available Hugging Face repositories. All code was written in Python 3.10.18, employing torchvision 0.16.2 for image transformations and Transformers 4.50.2 for loading ViT and ResNet checkpoints. Classical machine-learning utilities, such as cross-validation, stratified splitting, and computation of AUROC, balanced accuracy, F1-score, and regression metrics, were implemented using scikit-learn 1.7.0. Data handling and preprocessing utilized NumPy 1.26.4 and pandas 2.3.0. Visualization and figure generation were performed using Matplotlib 3.10.3 and Seaborn 0.13.2. Additional Python dependencies used in this study are listed in the Reporting Summary.

**Evaluation and statistical analysis**

The normality of all data distributions was assessed using the Shapiro–Wilk test. For inter-model performance comparisons, either a two-sided independent t-test or a Mann–Whitney U test was applied depending on the normality of the data, under the null hypothesis that no significant difference existed between SKELEX and the baseline models. For analyses involving zero-shot abnormality maps, the Mann–Whitney U test was used to compare the mean squared errors between the original and reconstructed radiographs of normal and abnormal cases. A P-value < 0.05 was considered statistically significant. All statistical analyses were performed using SciPy library (version 1.5.4).

**Data availability**

All publicly available datasets analyzed in this study can be accessed through their respective data portals: MURA (https://aimi.stanford.edu/datasets/mura-msk-xrays), GRAZPEDWRI-DX (https://doi.org/10.6084/m9.figshare.14825193), RSNA Bone Age (https://www.rsna.org/rsnai/ai-image-challenge/rsna-pediatric-bone-age-challenge-2017), FRACATLAS (https://figshare.com/articles/dataset/The_dataset/22363012?file=43283628), Pes Planus Dataset (https://doi.org/10.3390/diagnostics13091662), OAI (https://data.mendeley.com/datasets/56rmx5bjcr/1), BTXRD (https://doi.org/10.6084/m9.figshare.27865398)

The SNUH-1M and SNUH-BoneTumor datasets consist of retrospectively collected, de-identified radiographs and are subject to institutional data-use and privacy regulations. Access may be granted upon reasonable request. All requests will be reviewed on a case-by-case basis to ensure compliance with institutional policies, intellectual property considerations, and patient privacy protections.

**Code availability**

The authors will release code for performing downstream evaluation using the pretrained encoder of this work upon publication.

**Acknowledgments**

This work was supported by a grant from the MD-PhD/Medical Scientist Training Program through the Korea Health Industry Development Institute (KHIDI), funded by the Ministry of Health and Welfare, Republic of Korea; the Korea-US Collaborative Research Fund(KUCRF), funded by the Ministry of Science and ICT and Ministry of Health & Welfare, Republic of Korea (grant number: RS-2024-00468338); the Industrial Strategic Technology Development Program funded By the Ministry of Trade Industry & Energy (MOTIE) of Republic of Korea (RS-2024-00508416); the BK21 FOUR program of the Education and Research Program for Future ICT Pioneers, Seoul National University in 2025

**Author contributions**

S.Kim, S.L., I.H., S.Kwon conceptualized the project. S.Kim and S.L. performed the experiments and data analysis and wrote the initial manuscript. K.S., J.N., S.Ko, D.K., W.H. provided critical feedback and contributed to the interpretation of the results. H.-S.K., Y.K., and I.H. provided clinical resources and coordinated data acquisition. M.K. developed the web application. All authors contributed to manuscript review and editing. I.H. and S.Kwon provided overall supervision of the study.

**Competing interests**

Authors declare that they have no competing interests.

# Figures

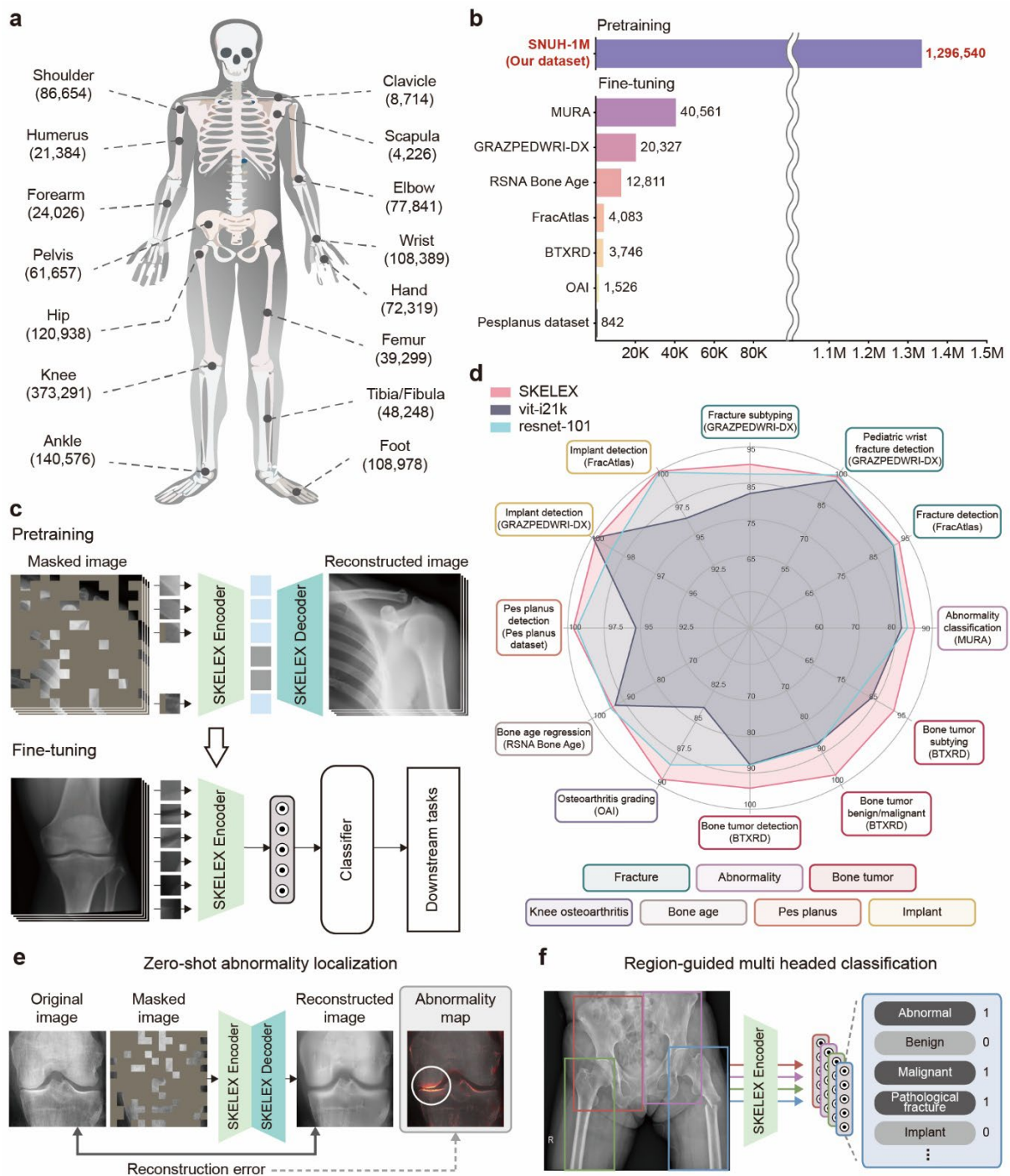

**Figure 1. Overview of SKELEX.**

SKELEX represents a large-scale foundation model for musculoskeletal (MSK) radiographs, demonstrating strong performance across 12 diagnostic tasks, with capabilities for zero-shot abnormality localization and region-level interpretable prediction. a, Distribution of the SNUH-1M dataset, comprising more than 1.2 million radiographs across 15 body parts. b, SKELEX was pretrained on SNUH-1M, a dataset approximately 30 times larger than the largest publicly available MSK dataset, and subsequently fine-tuned on multiple open datasets, including MURA, GRAZPEDWRI-DX, RSNA Bone Age, FracAtlas, OAI, and Pes Planus. c, SKELEX was pretrained using a masked autoencoding strategy, in which randomly masked image regions were reconstructed to learn contextual visual representations, followed by fine-tuning for diverse diagnostic tasks. d, SKELEX outperformed other pretrained encoders across 12 clinical benchmarks in musculoskeletal radiography. e, The model achieved zero-shot abnormality localization by generating reconstruction error maps without explicit localization supervision. f, Region-guided multi-headed model fine-tuned from SKELEX to generate region-interpretable predictions for bone tumors, fractures, and implants.

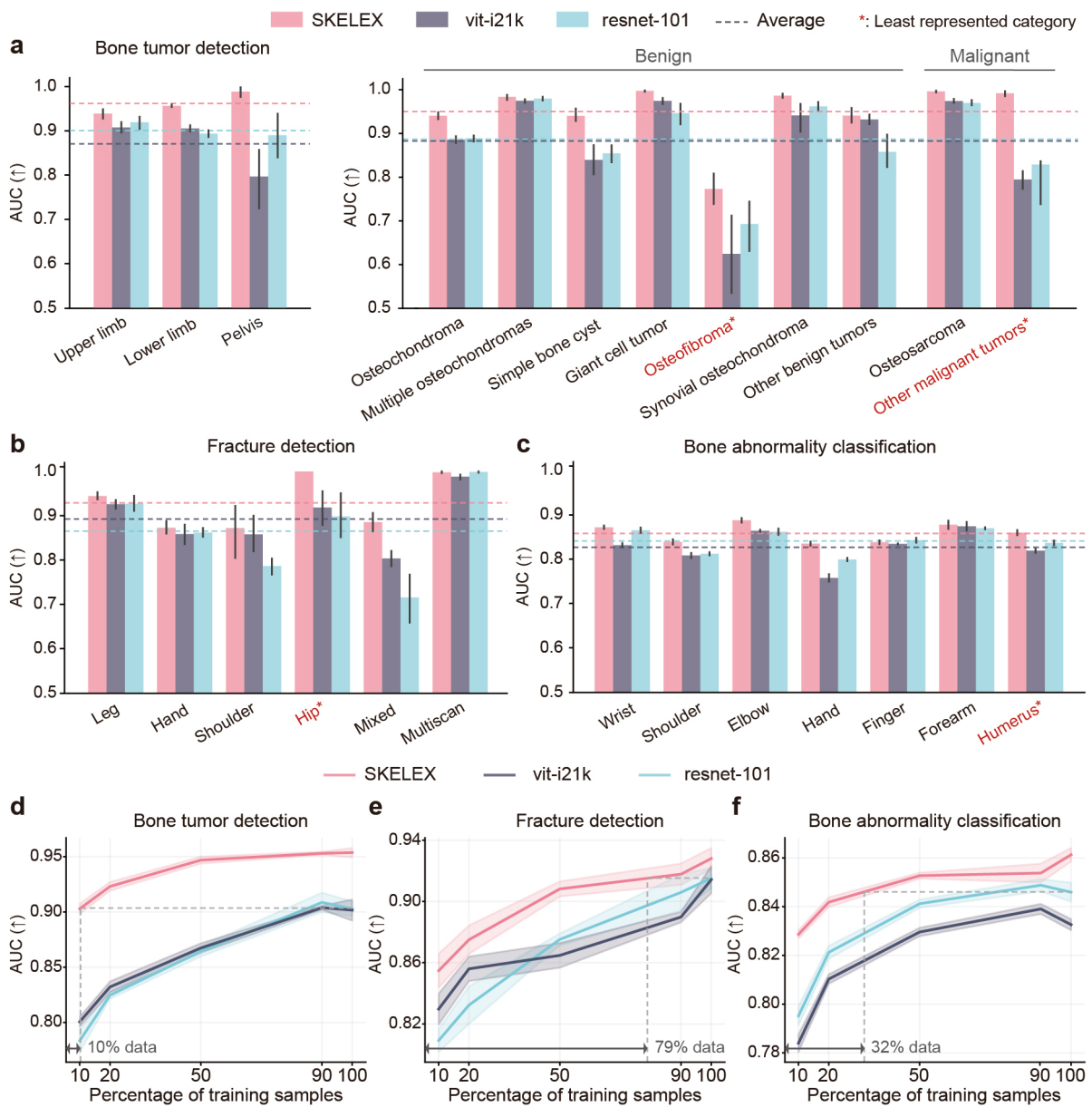

**Figure 2. Performance across musculoskeletal disease diagnostic tasks.**

SKELEX demonstrated robust performance across multiple musculoskeletal disease classification tasks and showed strong label efficiency. a, Bone tumor classification performance of SKELEX and baseline models across different anatomical regions and bone tumor subtypes. b, c, Fracture (b) and bone abnormality (c) classification performance of SKELEX and baseline models across various anatomical regions. Dashed lines indicate the mean performance of each model across all categories. All values represent the area under the receiver operating characteristic curve (AUROC), and error bars indicate 95% confidence intervals. Numbers shown below each bar indicate the number of training samples for each category. Red text with an asterisk (*) denotes the category with the smallest sample size in the dataset. d–f, Label efficiency analyses showing model performance with varying proportions of labeled training data, demonstrating the amount of data required to achieve target performance levels. d, Bone tumor classification. e, Fracture classification. f, Bone abnormality classification. The grey dashed line marks the fraction of labeled training data at which our model attains the maximum performance reached by baseline models using the entire training set.

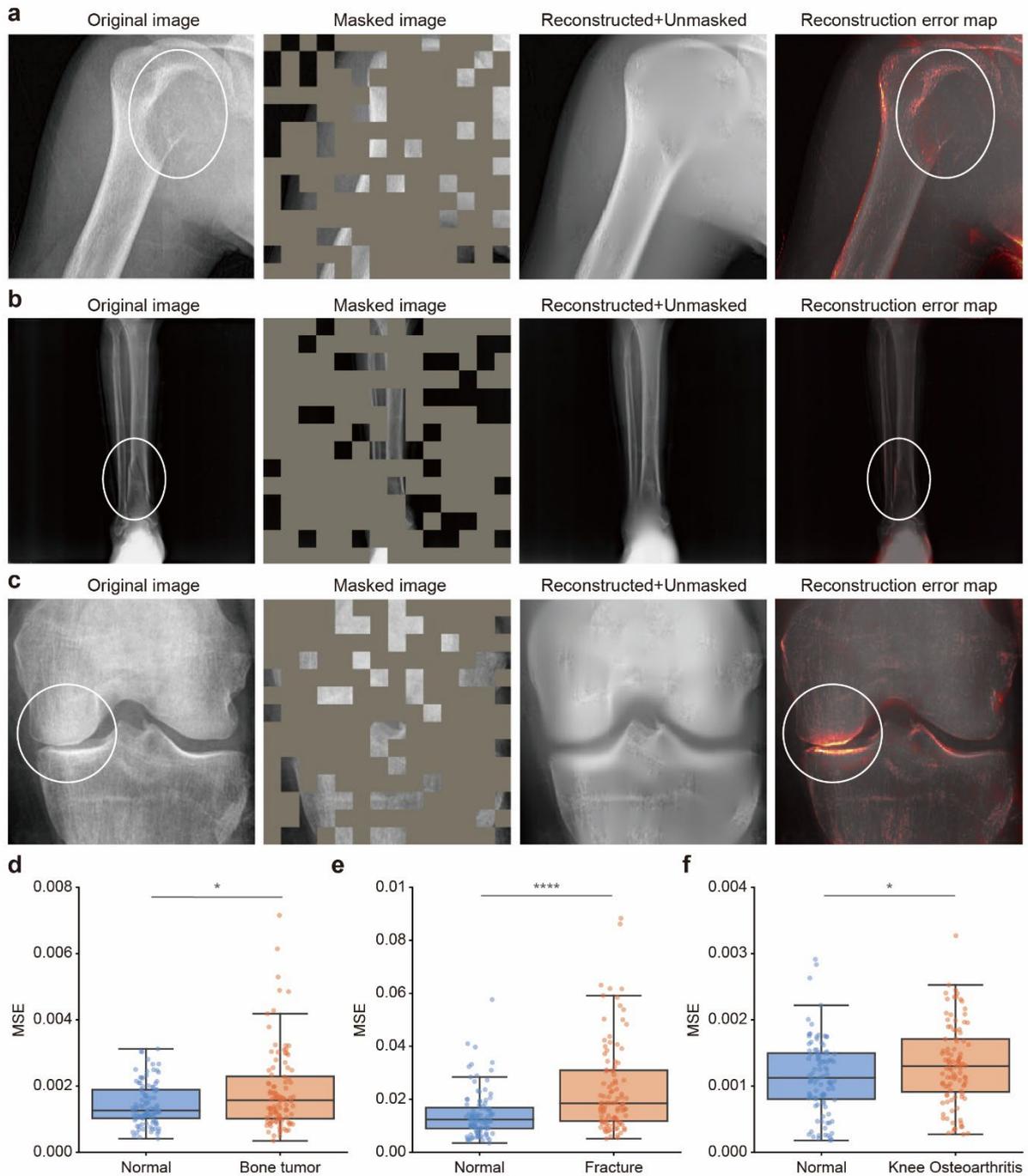

**Figure 3. Zero-shot error map generation with SKELEX.**

SKELEX produces error maps that localize abnormal regions without any task-specific training. a–c, Error maps generated from the BTXRD, FracAtlas, and OAI datasets, corresponding to bone tumor (a), fracture (b), and osteoarthritis (c) cases, respectively. The maps highlight regions of abnormality for each condition. d–f, Quantitative comparisons of mean reconstruction error values between abnormal and normal radiographs in the BTXRD (d), FracAtlas (e), and OAI (f) datasets (*$P < 0.05$, **$P < 0.01$, ***$P < 0.001$, ****$P < 0.0001$; Mann-Whitney U test).

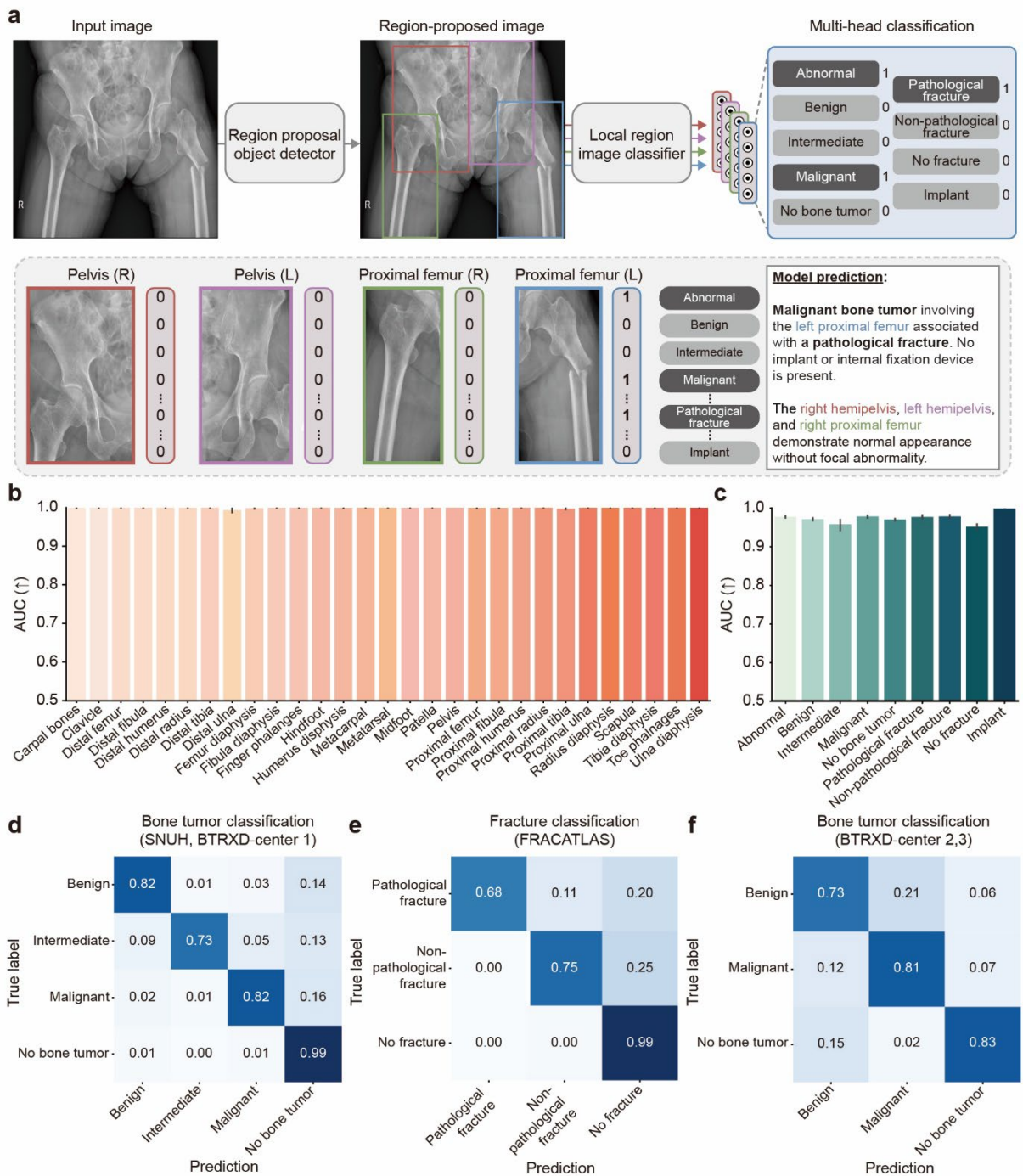

**Figure 4. Interpretable region-guided multi-headed bone tumor classifier with auxiliary heads using SKELEX**

SKELEX can be fine-tuned into an interpretable, multi-headed bone tumor classifier with auxiliary heads including fracture and implant detection. a, Schematic illustration of the two-stage framework comprising anatomical region localization followed by region-specific inference for multi-headed musculoskeletal abnormality classification. b, c, 5-fold cross-validation performance of the multi-headed local classifier for anatomical localization (b) and for abnormality, bone tumor, fracture, and implant classification (c). All values represent the area under the receiver operating characteristic curve (AUROC), and error bars indicate 95% confidence intervals. d, e, Confusion matrices showing internal test performance for bone tumor (d) and fracture (e) classification. f, Confusion matrix of external validation on independent datasets (BTXRD-Centers 2 and 3, sourced from Radiopaedia and MedPix).

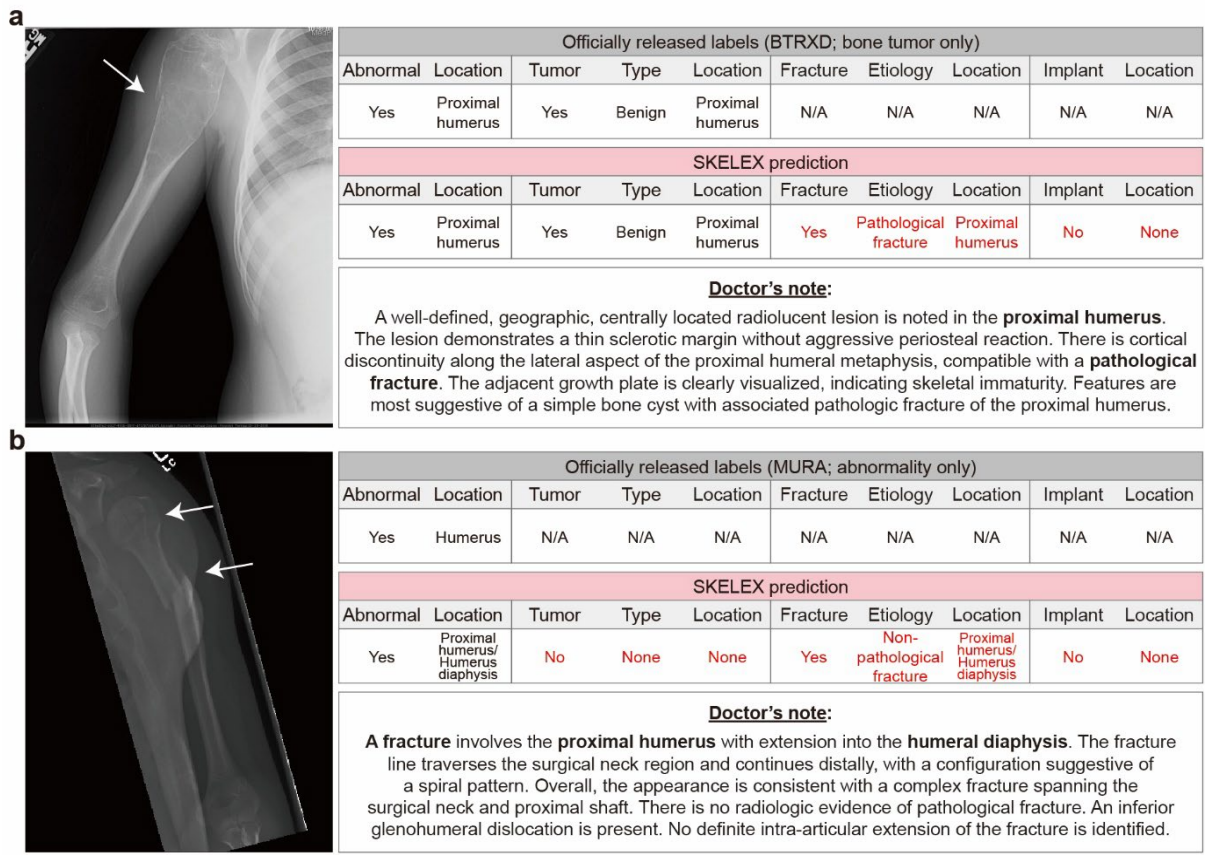

**Figure 5. Model interpretive capability beyond dataset-provided labels.**

The region-guided multi-headed bone tumor classifier demonstrated interpretive capacity that exceeded the label available in publicly released datasets. a, In the BTXRD dataset, which lacks explicit fracture annotations despite containing bone tumors with fracture cases, the model correctly identified both bone tumors and fractures, thereby extending beyond the official dataset labels and aligning with clinical reports. b, In the MURA dataset, which provides only binary abnormality labels without specifying the underlying pathology, the model successfully inferred detailed findings including bone tumor, fracture, implant, and anatomical location. Predictions by SKELEX not present in the original dataset labels are highlighted in red, and the accompanying clinical notes were written by a board-certified orthopedic surgeon.

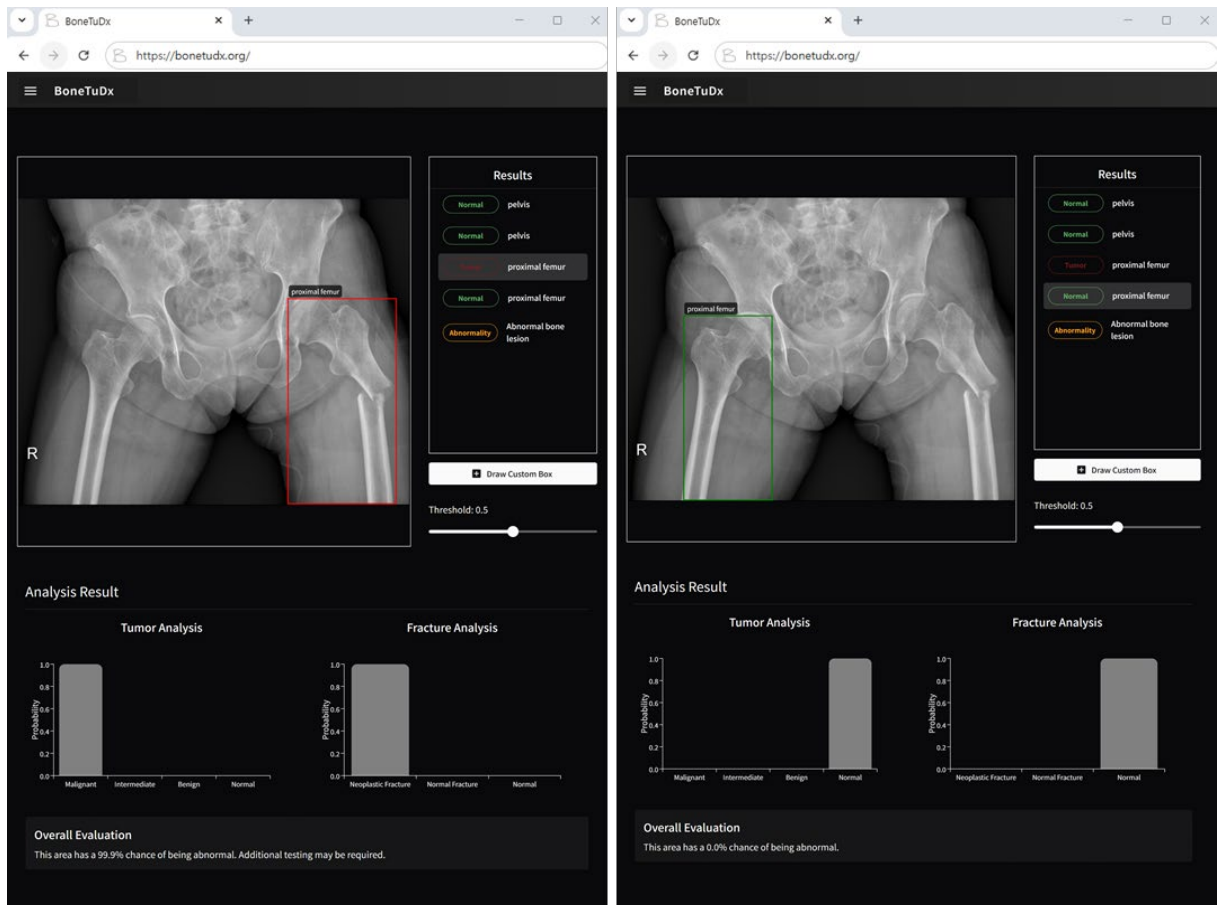

**Figure 6. Web-based diagnostic interface demonstrating the clinical applicability of the SKELEX-derived model**

A publicly accessible web interface allows users to upload musculoskeletal radiographs and automatically estimate the probabilities of bone tumor and fracture. Anatomical regions identified by the region-proposal network are visualized with bounding boxes, and corresponding classification probabilities are displayed for each region. Users may interactively adjust the decision threshold to explore how predictions change across operating points.

classification compendium—2018. *J. Orthop. Trauma* **32**, S1–S10 (2018).

# Supplementary Information: A generalizable large-scale foundation model for musculoskeletal radiographs

## Contents

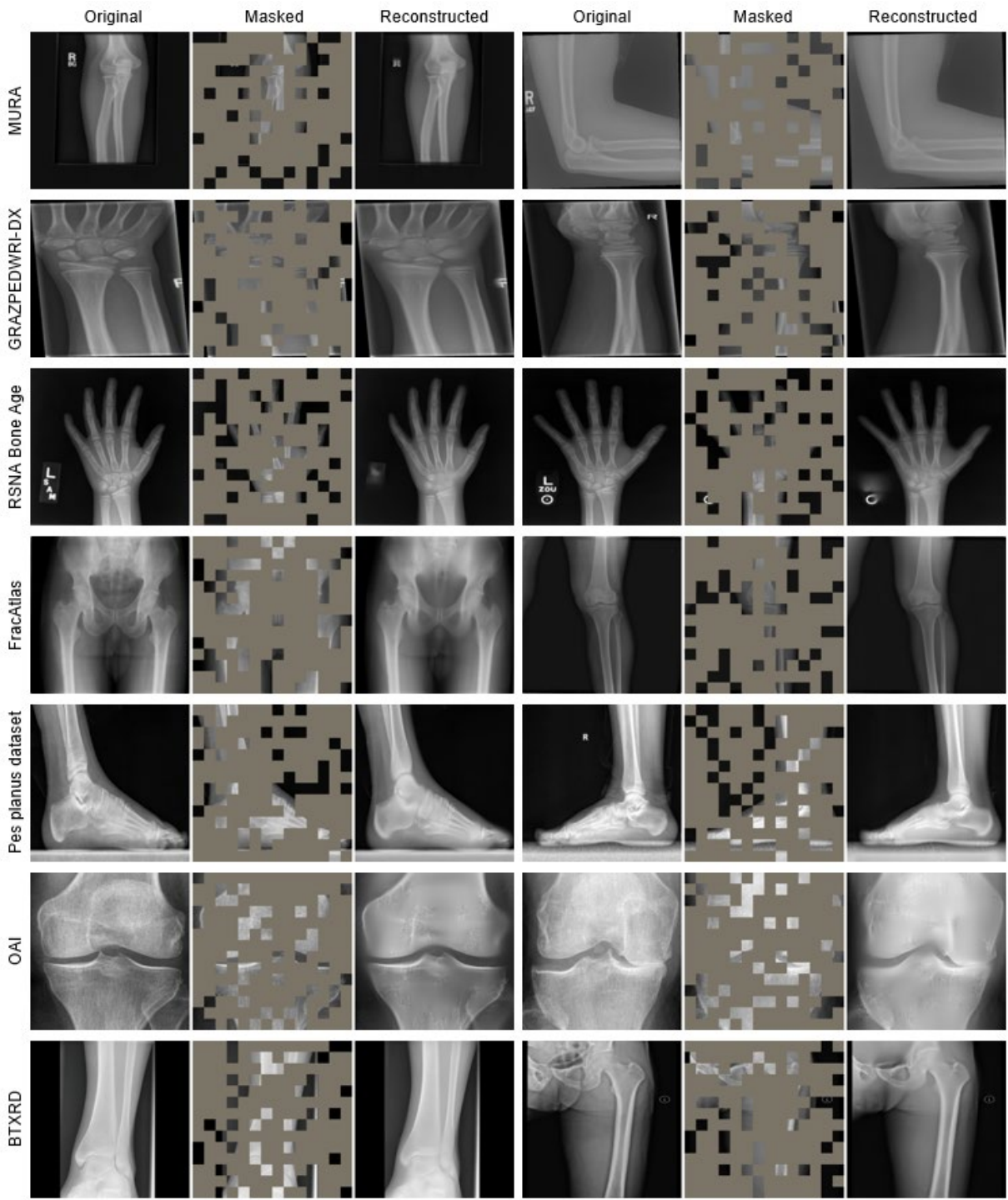

**Supplementary Figure 1. SKELEX reconstructs musculoskeletal radiographs from diverse datasets.**

Radiographs from diverse anatomical regions and datasets are accurately reconstructed following random masking. Joint structures and key anatomical landmarks are accurately recovered. Each triplet displays the original radiograph (left), the randomly masked input (middle), and the output image (right), where masked regions are reconstructed by SKELEX and unmasked regions are retained from the original image.

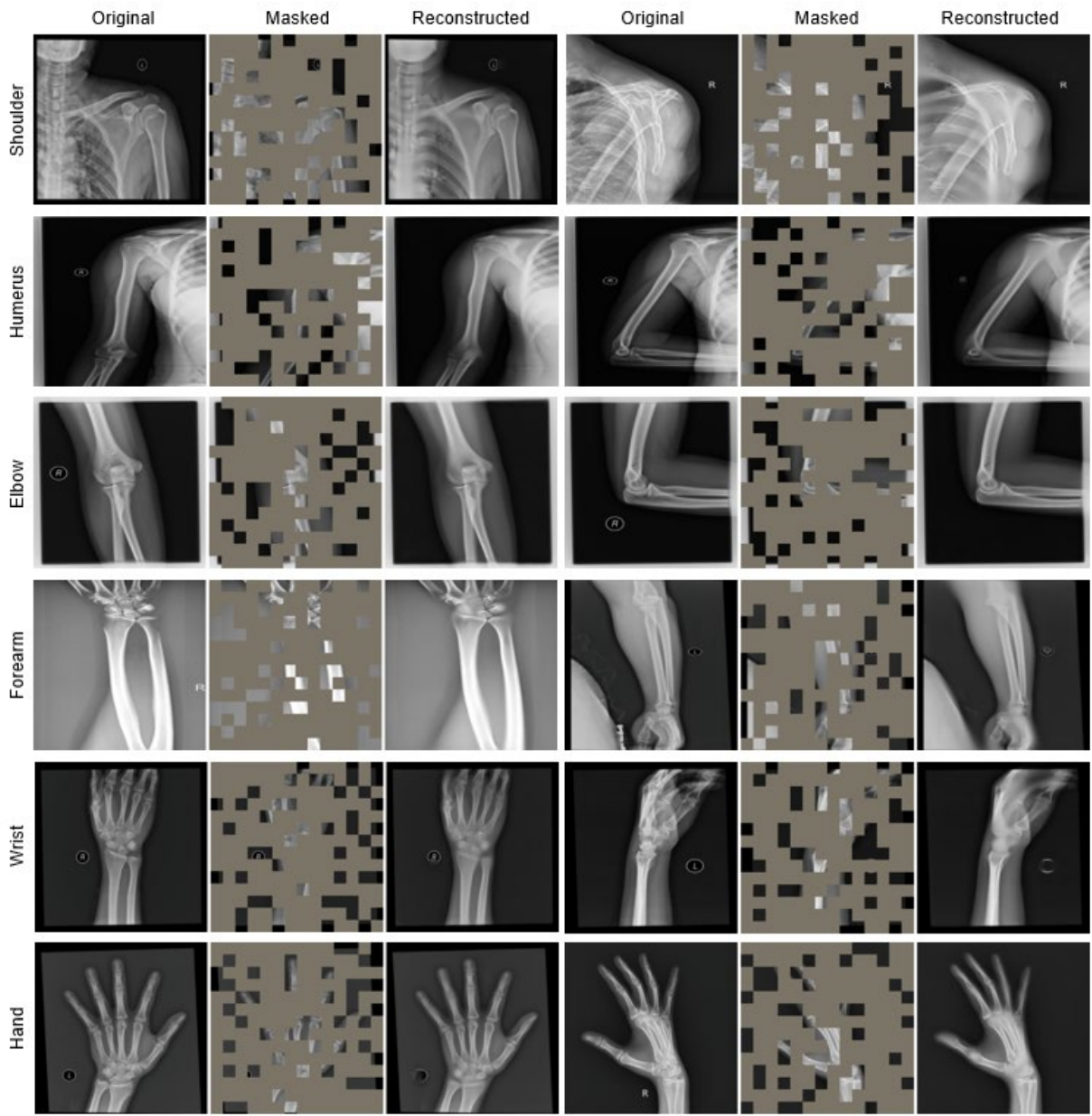

**Supplementary Figure 2. SKELEX reconstructs musculoskeletal radiographs across diverse upper-extremity anatomical regions and projection views.**

Across all upper-extremity anatomical regions, radiographs from diverse projection views are accurately reconstructed. Each triplet displays the original radiograph (left), the randomly masked input (middle), and the output image (right), where masked regions are reconstructed by SKELEX and unmasked regions are retained from the original image. All images represent normal cases from the BTXRD[36] dataset and were not included in the training data used to develop SKELEX.

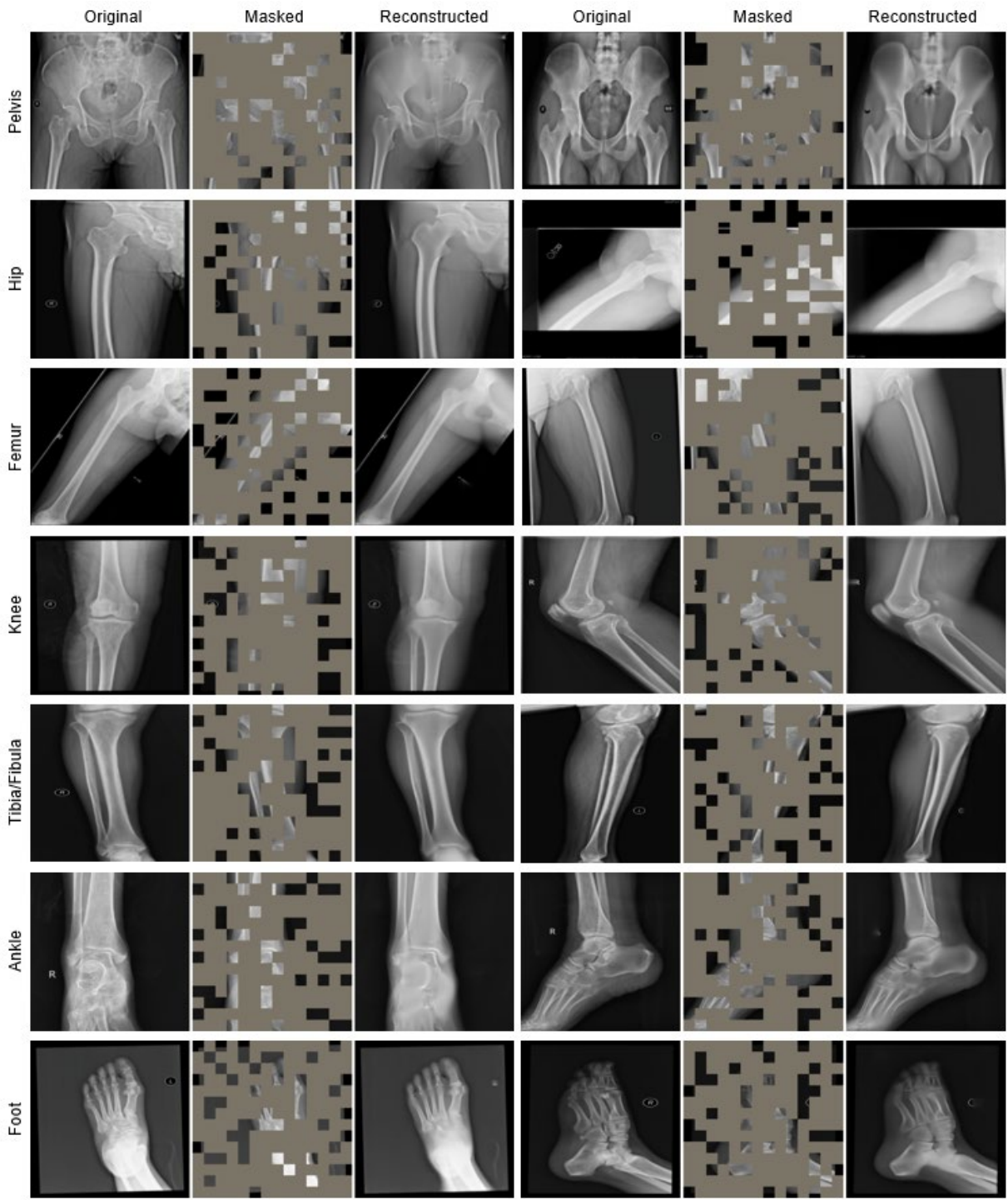

**Supplementary Figure 3. SKELEX reconstructs musculoskeletal radiographs across diverse lower-extremity anatomical regions and projection views.**

Across all lower-extremity anatomical regions, radiographs from diverse projection views are accurately reconstructed. Each triplet displays the original radiograph (left), the randomly masked input (middle), and the output image (right), where masked regions are reconstructed by SKELEX and unmasked regions are retained from the original image. All images represent normal cases from the BTXRD[36] dataset and were not included in the training data used to develop SKELEX.

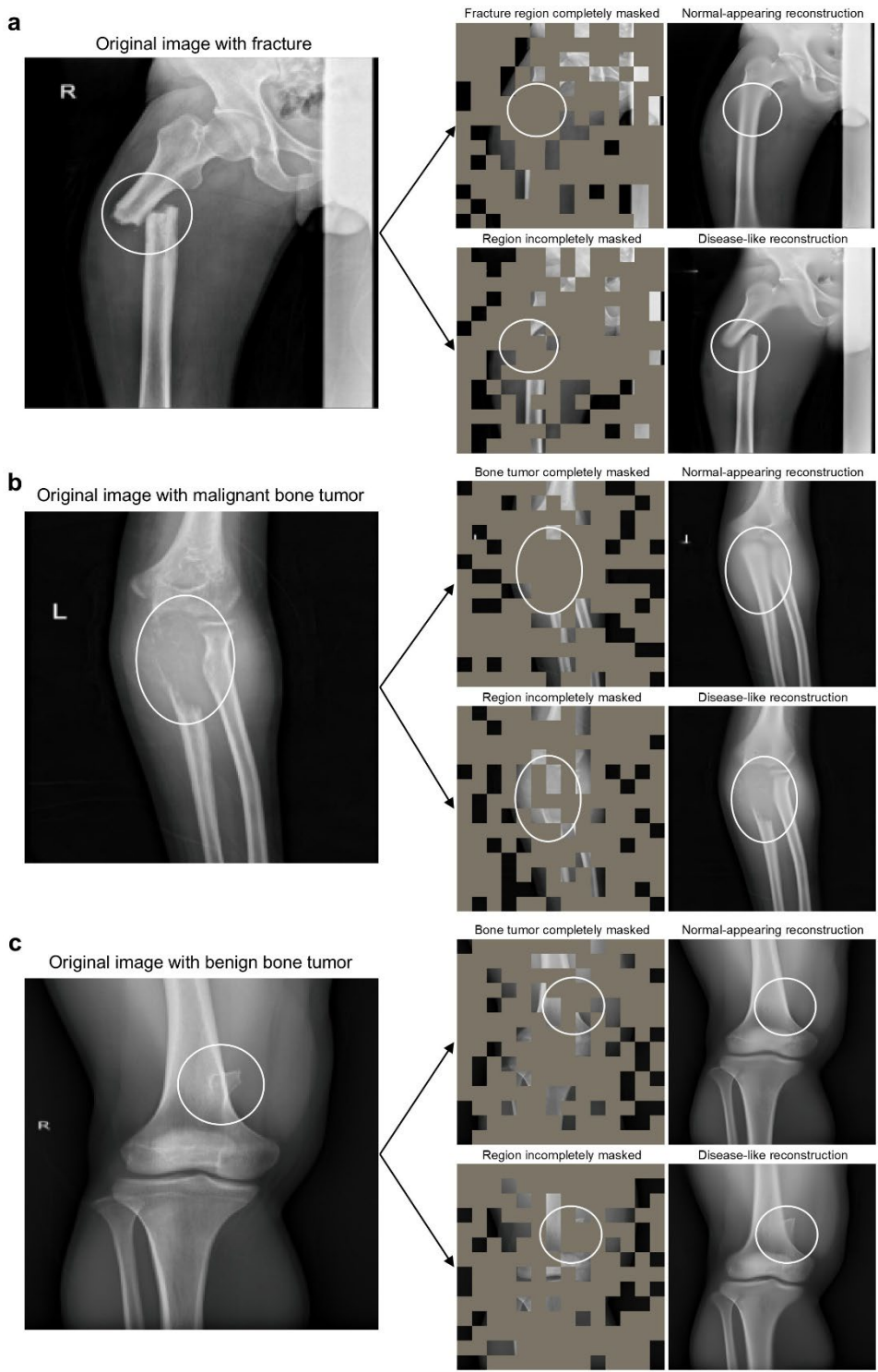

**Supplementary Figure 4. Mask coverage influences reconstruction of pathologic regions.** SKELEX reconstructs radiographs as either normal- or disease-like depending on the degree to which pathologic regions are masked. Complete masking of fractures or bone tumors produces normal-appearing reconstructions, whereas partial masking preserves abnormal features. **a**, Fracture. **b**, Malignant bone tumor. **c**, Benign bone tumor. All images are from the BTXRD[36] dataset and were not included in the training data used to develop SKELEX.

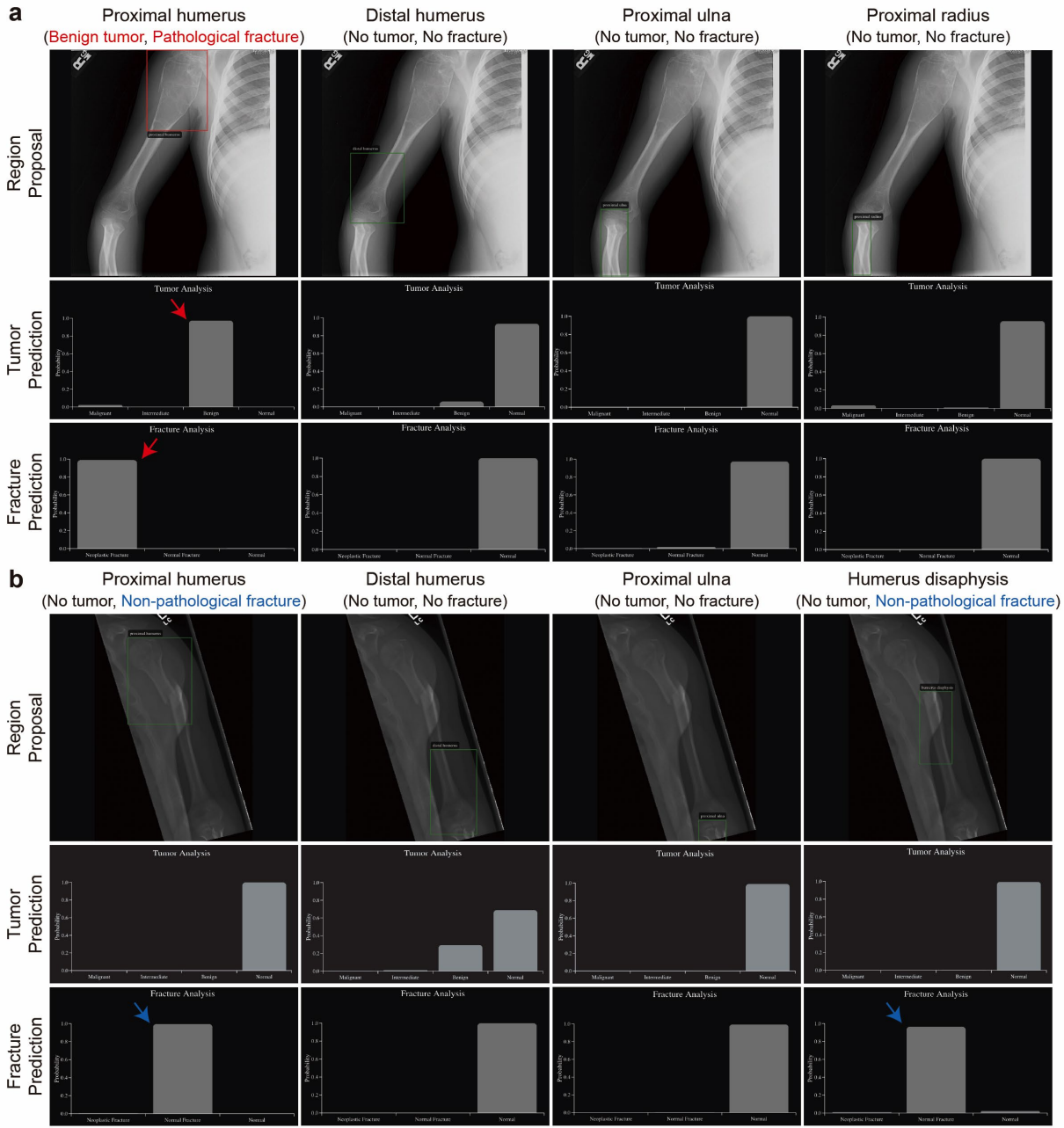

**Supplementary Figure 5. Detailed output of region-guided multi-headed bone tumor classification.**

The region-guided multi-headed classifier identifies anatomical regions and provides detailed predictions for each, including bone tumor presence, tumor subtype (malignant, intermediate, benign), fracture presence, and fracture category (neoplastic pathologic vs. non-neoplastic). **a,** Proximal humeral benign bone tumor with a neoplastic pathologic fracture. The model correctly identifies a benign tumor with pathological fracture in the proximal humerus, while classifying the distal humerus, proximal ulna, and proximal radius as free of tumor and fracture. The original BTXRD dataset did not include fracture annotations for this case despite the presence of a pathological fracture (see Fig. 5 for details). **b,** Proximal humeral and humeral diaphyseal non-neoplastic fractures. The model correctly predicts non-neoplastic fractures in the proximal humerus and humeral diaphysis, while classifying the distal humerus and proximal ulna as having no evidence of tumor or fracture. This radiograph originated from the MURA dataset, where only a binary abnormality label was provided, without specification of tumor or fracture type (see Fig. 5 for details).

| Body Part | Num. Radiographs |
|---|---:|
| Knee | 373,291 |
| Ankle | 140,576 |
| Hip | 120,938 |
| Foot | 108,978 |
| Wrist | 108,389 |
| Shoulder | 86,654 |
| Elbow | 77,841 |
| Hand | 72,319 |
| Pelvis | 61,657 |
| Tibia Fibula | 48,248 |
| Femur | 39,299 |
| Forearm | 24,026 |
| Humerus | 21,384 |
| Clavicle | 8,714 |
| Scapula | 4,226 |
| **Total** | **1,296,540** |

Supplementary Table 1: **Distribution of major body parts in the SNUH-1M dataset.** SNUH-1M is a pretraining dataset comprising of 1,296,540 diagnostic musculoskeletal radiographs of the extremities, spanning 15 major body parts collected at Seoul National University Hospital (SNUH).

| Category | Report Diagnostic Keywords | n |
|---|---|---|
| **Congenital & Developmental** | | |
| | *Accessory bone / variant*: accessory navicular | 9,381 |
| | os subfibulare | 1,762 |
| | bipartite patella | 528 |
| | patella alta | 223 |
| | **Subtotal** | **11,894** |
| | *Foot deformity*: pes planus | 4,856 |
| | clubfoot | 1,590 |
| | pes planovalgus | 1,071 |
| | flatfoot | 704 |
| | equinocavovarus | 76 |
| | calcaneovalgus | 28 |
| | **Subtotal** | **8,325** |
| | *Digit anomaly*: polydactyly | 3,633 |
| | syndactyly | 740 |
| | oligodactyly | 306 |
| | acrosyndactyly | 131 |
| | symbrachydactyly | 38 |
| | oligosyndactyly | 37 |
| | **Subtotal** | **4,885** |
| | *Hip dysplasia*: ddh | 2,428 |
| | **Subtotal** | **2,428** |
| | *Limb length discrepency*: lld | 1,150 |
| | leg length discrepancy | 659 |
| | limb length discrepancy | 585 |
| | **Subtotal** | **2,394** |
| | *Growth modulation*: epiphysiodesis | 893 |
| | hemiepiphysiodesis | 704 |
| | **Subtotal** | **1,597** |
| | *Skeletal dysplasia*: osteogenesis imperfecta | 266 |
| | achondroplasia | 118 |
| | spondyloepiphyseal dysplasia | 112 |
| | pseudoachondroplasia | 90 |
| | chondrodysplasia | 44 |
| | hypochondroplasia | 30 |
| | **Subtotal** | **660** |
| | *Digit deformity*: brachymetatarsia | 262 |
| | clinodactyly | 226 |
| | brachymetacarpia | 134 |
| | camptodactyly | 117 |
| | **Subtotal** | **739** |
| | *Limb deficiency*: hemimmelia | 170 |
| | hemihypotrophy | 80 |
| | hemimelia | 67 |
| | **Subtotal** | **317** |
| | *Wrist deformity*: madelung | 134 |
| | **Subtotal** | **134** |
| | **Category Total** | **33,373** |
| **Deformity** | | |
| | *Spinal deformity*: scoliosis | 12,217 |
| | **Subtotal** | **12,217** |
| | *Adult foot deformity*: hallux valgus | 18,873 |

| Category | Report Diagnostic Keywords | n |
|---|---|---|
| | **Subtotal** | **18,873** |
| | *Knee alignment deformity*: genu varum | 10,042 |
| |   genu varus | 3,854 |
| |   valgus type | 1,370 |
| |   deformity knee | 1,037 |
| |   recurvatum | 37 |
| | **Subtotal** | **16,340** |
| | *Hip deformity*: coxa valga | 1,370 |
| | **Subtotal** | **1,370** |
| | *Finger deformity*: boutonniere | 100 |
| | **Subtotal** | **100** |
| | **Category Total** | **48,900** |
| **Degenerative** | | |
| | *Osteoarthritis*: oa | 211,089 |
| |   degenerative | 18,649 |
| |   joint space narrowing | 13,853 |
| |   kl grade | 8,187 |
| |   kl score | 4,510 |
| | **Subtotal** | **256,288** |
| | *Degenerative spine disease*: spondylosis | 62,527 |
| |   disc space narrowing | 39,094 |
| | **Subtotal** | **101,621** |
| | *Subacromial impingement*: subacromial spur | 18,846 |
| |   spur ac | 753 |
| | **Subtotal** | **19,599** |
| | *Spinal instability*: spondylolisthesis | 13,439 |
| |   retrolisthesis | 8,491 |
| | **Subtotal** | **21,930** |
| | *Shoulder degeneration*: decreased acromiohumeral distance | 538 |
| |   rotator cuff arthropathy | 471 |
| | **Subtotal** | **1,009** |
| | **Category Total** | **400,447** |
| **Infectious** | | |
| | *Bone infection*: osteomyelitis | 770 |
| | **Subtotal** | **770** |
| | *Joint infection*: infectious arthritis | 724 |
| |   septic arthritis | 558 |
| | **Subtotal** | **1,282** |
| | *Granulomatous disease*: granulomas | 460 |
| | **Subtotal** | **460** |
| | *Soft tissue / bone abscess*: abscess | 331 |
| | **Subtotal** | **331** |
| | *Infectious paralysis sequelae*: poliomyelitis | 141 |
| | **Subtotal** | **141** |
| | *Parasitic infection*: cysticercosis | 127 |
| | **Subtotal** | **127** |
| | **Category Total** | **3,111** |
| **Inflammatory & Autoimmune** | | |
| | *Rheumatoid arthritis*: ra | 5,451 |
| |   rheumatoid | 420 |

| Category | Report Diagnostic Keywords | n |
|---|---|---:|
| | **Subtotal** | **5,871** |
| | *CPPD / pseudogout*: chondrocalcinosis | 2,998 |
| |    cppd | 340 |
| |    pseudogout | 29 |
| | **Subtotal** | **3,367** |
| | *Inflammatory arthritis*: sacroiliitis | 2,378 |
| | **Subtotal** | **2,378** |
| | *Gouty arthritis*: gouty | 378 |
| | **Subtotal** | **378** |
| | *Inflammatory myopathy*: dermatomyositis | 92 |
| | **Subtotal** | **92** |
| | **Category Total** | **12,086** |
| **Metabolic & Endocrine** | | |
| | *Low bone density*: osteopenia | 70,262 |
| | **Subtotal** | **70,262** |
| | *Low bone density / fragility*: osteoporosis | 792 |
| | **Subtotal** | **792** |
| | *Defective mineralization*: rickets | 634 |
| |    hypophosphatemic | 103 |
| | **Subtotal** | **737** |
| | *Medication-related bone change*: bisphosphonate | 88 |
| | **Subtotal** | **88** |
| | *Marble bone disease*: osteopetrosis | 87 |
| | **Subtotal** | **87** |
| | *Hyperparathyroid bone disease*: renal osteodystrophy | 49 |
| |    hyperparathyroidism | 29 |
| |    brown tumors | 3 |
| | **Subtotal** | **81** |
| | *Paget's disease*: paget | 36 |
| | **Subtotal** | **36** |
| | *Endocrine overgrowth*: acromegaly | 9 |
| | **Subtotal** | **9** |
| | **Category Total** | **72,092** |
| **Neoplastic** | | |
| | *Chondrogenic tumors*: osteochondromatosis | 4,545 |
| |    osteochondroma | 3,960 |
| |    enchondroma | 1,757 |
| |    chondrosarcoma | 1,250 |
| |    enchondromatosis | 633 |
| |    chondroblastoma | 329 |
| |    enchondromas | 108 |
| |    bpop | 97 |
| |    maffucci | 63 |
| |    metachondromatosis | 29 |
| | **Subtotal** | **12,771** |
| | *Osteogenic tumor*: osteosarcoma | 3,099 |
| |    bone island | 2,349 |
| |    osteoid osteoma | 313 |
| | **Subtotal** | **5,761** |
| | *Other mesenchymal tumors of bone*: fibrous dysplasia | 2,186 |

| Category | Report Diagnostic Keywords | n |
|---|---|---|
| | intraosseous lipoma | 914 |
| | polyostotic fibrous dysplasia | 675 |
| | simple bone cyst | 532 |
| | osteofibrous dysplasia | 216 |
| | liposclerosing myxofibrous tumor | 77 |
| | ofd | 29 |
| | lsmft | 6 |
| | **Subtotal** | **4,635** |
| | *Osteoclastic giant cell rich tumor*: nof | 944 |
| | nonossifying fibroma | 177 |
| | giant cell tumor | 66 |
| | **Subtotal** | **1,187** |
| | *General malignant tumor*: sarcoma | 700 |
| | malignant | 413 |
| | malignancy | 96 |
| | **Subtotal** | **1,209** |
| | *Vascular tumors*: hemangioma | 641 |
| | hemangiomas | 37 |
| | **Subtotal** | **678** |
| | *Hematopoietic neoplasm*: lch | 321 |
| | plasmacytoma | 161 |
| | lymphoma | 131 |
| | **Subtotal** | **613** |
| | *Pigmented villonodular synovitis*: pvns | 264 |
| | **Subtotal** | **264** |
| | *Undifferentiated small round cell sarcoma*: ewing | 246 |
| | **Subtotal** | **246** |
| | *Liposarcoma*: liposarcoma | 143 |
| | **Subtotal** | **143** |
| | *Fibrogenic tumor*: fibromatosis | 140 |
| | fibrosarcoma | 32 |
| | **Subtotal** | **172** |
| | *Melanoma*: melanoma | 73 |
| | **Subtotal** | **73** |
| | *Lymphatic malformation*: lymphangioma | 29 |
| | **Subtotal** | **29** |
| | **Category Total** | **27,781** |
| **Trauma & Post-trauma & Postoperative** | | |
| | *General fracture*: fracture | 204,574 |
| | fx | 14,464 |
| | **Subtotal** | **219,038** |
| | *Postoperative / fixation device*: postop | 127,765 |
| | tkra | 51,782 |
| | orif | 46,120 |
| | internal fixation | 12,823 |
| | prosthesis | 7,930 |
| | screw | 7,818 |
| | splint | 7,398 |
| | cast | 5,818 |
| | crif | 3,290 |
| | im nail | 3,198 |

| Category | Report Diagnostic Keywords | n |
|---|---|---|
| | arthroplasty | 2,726 |
| | external fixator | 2,072 |
| | **Subtotal** | **278,740** |
| | *Fracture healing*: callus formation | 18,615 |
| | **Subtotal** | **18,615** |
| | *Intra-articular body*: loose bodies | 12,091 |
| | **Subtotal** | **12,091** |
| | *Dislocation / Subluxation*: dislocation | 10,111 |
| | subluxation | 8,548 |
| | shoulder dislocation | 709 |
| | radial head dislocation | 471 |
| | **Subtotal** | **19,839** |
| | *Fracture healing complication*: nonunion | 5,039 |
| | malunion | 627 |
| | pseudoarthrosis | 603 |
| | non osseous union | 438 |
| | **Subtotal** | **6,707** |
| | *Fracture through diseased bone*: pathologic fracture | 3,600 |
| | **Subtotal** | **3,600** |
| | *Reactive bone change*: periosteal reaction | 2,204 |
| | **Subtotal** | **2,204** |
| | *Spinal instability*: spondylolytic spondylolisthesis | 2,014 |
| | **Subtotal** | **2,014** |
| | *Limb lengthening procedure*: distraction osteogenesis | 1,043 |
| | **Subtotal** | **1,043** |
| | *Tendon tear*: rotator cuff tear | 735 |
| | **Subtotal** | **735** |
| | *Joint instability*: anterior instability knee | 537 |
| | anterior drawer | 470 |
| | lateral instability ankle | 450 |
| | instability knee joint | 425 |
| | **Subtotal** | **1,882** |
| | *Shoulder instability lesion*: hill sachs lesion | 504 |
| | **Subtotal** | **504** |
| | *Pediatric hip disorder*: scfe | 473 |
| | slipped capital femoral epiphysis | 49 |
| | **Subtotal** | **522** |
| | *Collapse from AVN / trauma*: femoral head collapse | 470 |
| | **Subtotal** | **470** |
| | *Prosthesis complication*: periprosthetic osteolytic lesion | 424 |
| | **Subtotal** | **424** |
| | *Physeal fracture*: salter harris type | 413 |
| | **Subtotal** | **413** |
| | **Category Total** | **568,841** |
| **Vascular & Necrotic** | | |
| | *Avascular necrosis*: avn | 5,007 |
| | femoral head osteonecrosis | 271 |
| | **Subtotal** | **5,278** |
| | *Lunate AVN*: kienbock disease | 692 |

| Category | Report Diagnostic Keywords | n |
|---|---|---|
| | **Subtotal** | **692** |
| | *Metatarsal head AVN*: freiberg disease | 168 |
| | **Subtotal** | **168** |
| | *Vanishing bone disease*: gorham | 149 |
| | **Subtotal** | **149** |
| | *Bone infarct*: infarction | 73 |
| | **Subtotal** | **73** |
| | **Category Total** | **6,360** |
| **Radiographic Finding** | | |
| | *Joint effusion*: joint effusion | 32,560 |
| | **Subtotal** | **32,560** |
| | *Soft tissue change*: soft tissue swelling | 21,496 |
| |    soft tissue edema | 1,015 |
| | **Subtotal** | **22,511** |
| | *Lesion descriptor*: osteolytic lesion | 11,693 |
| |    osteolytic lesions | 1,597 |
| |    radiolucent lesion | 1,372 |
| |    osteoblastic | 1,039 |
| |    expansile osteolytic lesion | 755 |
| |    geographic osteolytic lesion | 649 |
| |    patchy | 472 |
| |    ggo | 456 |
| | **Subtotal** | **18,033** |
| | *Cystic lesion*: subchondral cyst | 1,862 |
| | **Subtotal** | **1,862** |
| | *Vascular calcification*: phleboliths | 223 |
| |    phlebolith | 217 |
| | **Subtotal** | **440** |
| | **Category Total** | **75,406** |
| **Miscellaneous** | | |
| | *Calcific tendinopathy*: calcific tendinitis | 5,714 |
| |    hadd | 495 |
| |    hydroxyapatite | 37 |
| | **Subtotal** | **6,246** |
| | *Soft tissue mineralization*: heterotopic ossification | 4,601 |
| |    heterotopic ossifications | 753 |
| |    soft tissue calcifications | 444 |
| |    soft tissue ossification | 394 |
| |    myositis ossificans | 76 |
| | **Subtotal** | **6,268** |
| | *Enthesopathy*: calcaneal spur | 4,442 |
| |    enthesopathy | 258 |
| | **Subtotal** | **4,700** |
| | *Popliteal cyst*: baker cyst | 291 |
| | **Subtotal** | **291** |
| | *Arterial calcification*: monckeberg | 255 |
| | **Subtotal** | **255** |
| | *Heel deformity*: haglund | 148 |
| | **Subtotal** | **148** |
| | *Sclerosing bone dysplasia*: melorheostosis | 119 |

| Category | Report Diagnostic Keywords | n |
|---|---|---|
| | **Subtotal** | **119** |
| | *Tendon pathology*: tendinopathy | 97 |
| | **Subtotal** | **97** |
| | *Bleeding disorder*: hemophilia | 33 |
| |     hemophilic arthropathy | 7 |
| | **Subtotal** | **40** |
| | **Category Total** | **18,164** |

Supplementary Table 2: **Keyword frequencies from radiograph reports in the SNUH-1M dataset.** This dataset includes 11 major diagnostic categories and 89 subcategories, encompassing more than 100 unique keywords that capture a broad spectrum of musculoskeletal conditions.

| Dataset | Release year | Images | Body parts covered | Sources |
|---|---|---|---|---|
| MURA | 2017 | 40,561 | Upper extremity | https://aimi.stanford.edu/datasets/mura-msk-xrays |
| GRAZPEDWRI-DX | 2022 | 20,327 | Wrist | https://doi.org/10.6084/m9.figshare.14825193 |
| RSNA Bone Age | 2017 | 12,611 | Hand | https://www.rsna.org/rsnai/ai-image-challenge/rsna-pediatric-bone-age-challenge-2017 |
| FRACATLAS | 2023 | 4,083 | Hand, Shoulder, Hip, Leg | https://doi.org/10.6084/m9.figshare.22363012 |
| Pes planus dataset | 2023 | 842 | Foot | https://doi.org/10.3390/diagnostics13091662 |
| OAI | 2018 | 8,260 | Knee | https://data.mendeley.com/datasets/56rmx5bjcr/1 |
| BTXRD | 2025 | 3,746 | Tibia, Femur, Fibula, Humerus, Hand, Foot, Radius, Hip bone, Ulna, Knee, Shoulder, Elbow, Ankle, Wrist | https://doi.org/10.6084/m9.figshare.27865398 |

Supplementary Table 3: **Summary of publicly available datasets used in this study.**

| Model | Balanced ACC | F1 score | AUROC |
|---|---|---|---|
| SKELEX | 0.503 ± 0.027 | 0.489 ± 0.023 | 0.912 ± 0.006 |
| Vit-L/I21K | 0.341 ± 0.035 | 0.359 ± 0.039 | 0.847 ± 0.012 |
| ResNet-101 | 0.430 ± 0.010 | 0.433 ± 0.009 | 0.890 ± 0.017 |

Supplementary Table 4: **Evaluation of pediatric wrist fracture subtyping (GRAZPEDWRI-DX).**

| Model | Balanced ACC | F1 score | AUROC |
|---|---|---|---|
| SKELEX | 0.950 ± 0.005 | 0.947 ± 0.005 | 0.979 ± 0.001 |
| Vit-L/I21K | 0.912 ± 0.008 | 0.904 ± 0.009 | 0.965 ± 0.003 |
| ResNet-101 | 0.953 ± 0.005 | 0.951 ± 0.005 | 0.986 ± 0.002 |

Supplementary Table 5: **Evaluation of pediatric wrist fracture detection (GRAZPEDWRI-DX).**

| Model | Balanced ACC | F1 score | AUROC |
|---|---|---|---|
| SKELEX | 0.852 ± 0.005 | 0.762 ± 0.012 | 0.928 ± 0.007 |
| Vit-L/I21K | 0.789 ± 0.025 | 0.670 ± 0.037 | 0.914 ± 0.009 |
| ResNet-101 | 0.800 ± 0.020 | 0.713 ± 0.025 | 0.915 ± 0.007 |

Supplementary Table 6: **Evaluation of general fracture detection (FRACATLAS).**

| Model | Balanced ACC | F1 score | AUROC |
|---|---|---|---|
| SKELEX | 0.801 ± 0.003 | 0.763 ± 0.004 | 0.861 ± 0.003 |
| Vit-L/I21K | 0.764 ± 0.001 | 0.715 ± 0.001 | 0.833 ± 0.002 |
| ResNet-101 | 0.781 ± 0.004 | 0.737 ± 0.005 | 0.846 ± 0.004 |

Supplementary Table 7: **Evaluation of abnormality classification (MURA).**

| Model | Balanced ACC | F1 score | AUROC |
|---|---|---|---|
| SKELEX | 0.586 ± 0.031 | 0.593 ± 0.031 | 0.915 ± 0.007 |
| Vit-L/I21K | 0.436 ± 0.033 | 0.439 ± 0.038 | 0.858 ± 0.011 |
| ResNet-101 | 0.404 ± 0.020 | 0.418 ± 0.026 | 0.828 ± 0.016 |

Supplementary Table 8: **Evaluation of bone tumor subtyping (BTXRD).**

| Model | Balanced ACC | F1 score | AUROC |
|---|---|---|---|
| SKELEX | 0.845 ± 0.021 | 0.794 ± 0.029 | 0.975 ± 0.006 |
| Vit-L/I21K | 0.717 ± 0.014 | 0.572 ± 0.024 | 0.895 ± 0.012 |
| ResNet-101 | 0.730 ± 0.023 | 0.598 ± 0.034 | 0.900 ± 0.020 |

Supplementary Table 9: **Evaluation of bone tumor benign versus malignant classification (BTXRD).**

| Model | Balanced ACC | F1 score | AUROC |
|---|---|---|---|
| SKELEX | 0.895 ± 0.011 | 0.894 ± 0.011 | 0.954 ± 0.004 |
| Vit-L/I21K | 0.830 ± 0.012 | 0.826 ± 0.013 | 0.902 ± 0.010 |
| ResNet-101 | 0.828 ± 0.015 | 0.824 ± 0.019 | 0.903 ± 0.007 |

Supplementary Table 10: **Evaluation of bone tumor detection (BTXRD).**

| Model | Balanced ACC | F1 score | AUROC |
|---|---|---|---|
| SKELEX | 0.674 ± 0.007 | 0.660 ± 0.010 | 0.897 ± 0.005 |
| Vit-L/I21K | 0.602 ± 0.013 | 0.606 ± 0.011 | 0.851 ± 0.009 |
| ResNet-101 | 0.663 ± 0.013 | 0.648 ± 0.010 | 0.887 ± 0.004 |

Supplementary Table 11: **Evaluation of knee osteoarthritis grading (OAI).**

| Model | MAE | RMSE |
|---|---|---|
| SKELEX | 10.26 ± 0.05 | 13.34 ± 0.08 |
| Vit-L/I21K | 11.71 ± 0.19 | 15.14 ± 0.27 |
| ResNet-101 | 9.35 ± 0.13 | 12.03 ± 0.15 |

Supplementary Table 12: **Evaluation of bone age regression (RSNA bone age).**

| Model | Balanced ACC | F1 score | AUROC |
|---|---|---|---|
| SKELEX | 0.988 ± 0.000 | 0.988 ± 0.000 | 0.997 ± 0.003 |
| Vit-L/I21K | 0.908 ± 0.032 | 0.905 ± 0.033 | 0.963 ± 0.020 |
| ResNet-101 | 0.976 ± 0.019 | 0.975 ± 0.019 | 0.995 ± 0.008 |

Supplementary Table 13: **Evaluation of pes planus detection (Pes Planus dataset).**

| Model | Balanced ACC | F1 score | AUROC |
|---|---|---|---|
| SKELEX | 0.994 ± 0.003 | 0.992 ± 0.005 | 0.999 ± 0.002 |
| Vit-L/I21K | 0.976 ± 0.005 | 0.976 ± 0.005 | 1.000 ± 0.001 |
| ResNet-101 | 0.978 ± 0.006 | 0.977 ± 0.006 | 0.994 ± 0.004 |

Supplementary Table 14: **Evaluation of implant detection in pediatric wrist radiographs (GRAZPEDWRI-DX).**

| Model | Balanced ACC | F1 score | AUROC |
|---|---|---|---|
| SKELEX | 0.990 ± 0.021 | 0.980 ± 0.040 | 1.000 ± 0.000 |
| Vit-L/I21K | 0.910 ± 0.020 | 0.901 ± 0.023 | 0.970 ± 0.011 |
| ResNet-101 | 0.950 ± 0.055 | 0.944 ± 0.065 | 0.999 ± 0.001 |

Supplementary Table 15: **Evaluation of implant detection in general fracture dataset (FracAtlas).**

| Anatomical Location | Num. Images |
| --- | --- |
| Distal Femur | 2,228 |
| Proximal Femur | 2,215 |
| Proximal Tibia | 1,642 |
| Proximal Fibula | 1,592 |
| Patella | 1,465 |
| Pelvis | 1,350 |
| Femur Diaphysis | 902 |
| Metacarpal | 815 |
| Metatarsal | 644 |
| Proximal Humerus | 605 |
| Clavicle | 531 |
| Distal Tibia | 492 |
| Finger Phalanges | 451 |
| Hindfoot | 434 |
| Distal Fibula | 429 |
| Proximal Radius | 398 |
| Distal Humerus | 383 |
| Scapula | 348 |
| Tibia Diaphysis | 334 |
| Proximal Ulna | 314 |
| Fibula Diaphysis | 292 |
| Humerus Diaphysis | 263 |
| Toe Phalanges | 240 |
| Distal Radius | 219 |
| Midfoot | 203 |
| Carpal Bones | 198 |
| Distal Ulna | 192 |
| Radius Diaphysis | 131 |
| Ulna Diaphysis | 109 |
| **Total** | **19,419** |

Supplementary Table 16: **Distribution of anatomical location in the SNUH-BoneTumor dataset.** SNUH-BoneTumor is a musculoskeletal bone tumor dataset comprising 19,757 cropped anatomical regions derived from 4,387 diagnostic musculoskeletal radiographs of the extremities collected at Seoul National University Hospital (SNUH).

| Config | Value |
| --- | --- |
| Encoder blocks | 24 |
| Embedding vector size | 1024 |
| Decoder blocks | 8 |
| Decoder vector size | 512 |
| Patch size | 16 |
| Hidden activation | GELU |
| Optimizer | AdamW |
| Learning rate | 7.5e-5 |
| Batch size | 128 |
| Weight decay | 0.05 |
| Optimizer momentum | $\beta_1, \beta_2 = 0.9, 0.999$ |
| Learning rate schedule | cosine decay |
| Warmup ratio | 0.05 |
| Epochs | 50 |
| Augmentation | RandomResizedCrop |
| Mask ratio | 0.75 |
| Normalized pixel loss | False |

Supplementary Table 17: **Hyperparameters of the masked autoencoder (MAE) used for ViT-L/16 pretraining.**

| Config | Value |
| --- | --- |
| Optimizer | AdamW |
| Learning rate | 5e-5 |
| Layerwise lr decay | 0.75 |
| Batch size | 64 |
| Weight decay | 0.05 |
| Optimizer momentum | $\beta_1, \beta_2 = 0.9, 0.999$ |
| Learning rate schedule | cosine decay |
| Warmup ratio | 0.1 |
| Epochs | 50 |

Supplementary Table 18: **Hyperparameters used for fine-tuning SKELEX and ViT-L/I21K.**

| Config | Value |
| --- | --- |
| Optimizer | AdamW |
| Learning rate | 5e-4 |
| Batch size | 64 |
| Weight decay | 1e-4 |
| Optimizer momentum | $\beta_1, \beta_2 = 0.9, 0.999$ |
| Learning rate schedule | cosine decay |
| Warmup ratio | 0.1 |
| Epochs | 50 |

Supplementary Table 19: **Hyperparameters used for fine-tuning ResNet-101.**

| AO Classification | Count |
|---|---:|
| 23r-M/2.1 | 2,174 |
| 23-M/2.1 | 674 |
| 23r-M/3.1 | 300 |
| 23r-M/3.1; 23u-M/2.1 | 230 |
| 23r-M/3.1; 23u-E/7 | 188 |
| 23r-M/2.1; 23u-E/7 | 156 |
| 23-M/3.1 | 133 |
| 23r-E/2.1 | 115 |
| 23r-E/2.1; 23u-E/7 | 91 |
| 23r-E/2.1; 23u-M/2.1 | 35 |
| **Total** | **4,096** |

Supplementary Table 20: **Top 10 pediatric wrist fracture subtypes (AO classification) in pediatric wrist fracture dataset (GRAZPEDWRI-DX).**